\documentclass{article}

\usepackage{arxiv}

\usepackage[utf8]{inputenc} 
\usepackage[T1]{fontenc}    
\usepackage{hyperref}       
\usepackage{url}            
\usepackage{booktabs}       
\usepackage{amsfonts}       
\usepackage{nicefrac}       
\usepackage{microtype}      
\usepackage{lipsum}		
\usepackage{graphicx}
\usepackage{natbib}
\usepackage{doi}
\usepackage{amsmath}
\usepackage{multirow}
\usepackage[table,xcdraw]{xcolor}

\title{From Geometric Mimicry to Comprehensive Generation: A Context-Informed Multimodal Diffusion Model for Urban Morphology Synthesis}

\date{Mar 12, 2026}	

\author{
\href{https://orcid.org/0009-0005-0484-1696}{\includegraphics[scale=0.06]{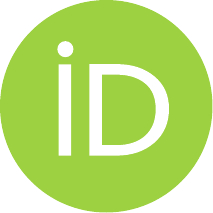}\hspace{1mm}Fangshuo Zhou}
\thanks{Fangshuo Zhou and Huaxia Li contributed equally to this work.}
\\
Zhejiang Agriculture and Forestry University\\
Hangzhou, 311300, China\\
\texttt{fangshuoz@stu.zafu.edu.cn}
\And
{Huaxia Li}
\footnotemark[1]
\\
Alibaba Group\\
Hangzhou, 311121, China\\
\texttt{lihx0610@gmail.com}
\And
\href{https://orcid.org/0000-0001-7635-7266}
{\includegraphics[scale=0.06]{orcid.pdf}\hspace{1mm}Liuchang Xu}
\thanks{Contact author: Liuchang Xu. Email: xuliuchang@zafu.edu.cn}
\\
Zhejiang Agriculture and Forestry University\\
Hangzhou, 311300, China\\
Zhejiang University\\
Hangzhou, 310058, China\\
\texttt{xuliuchang@zafu.edu.cn}
\And
{Rui Hu}
\\
Zhejiang University\\
Hangzhou, 310058, China\\
\And
{Sensen Wu}
\\
Zhejiang University\\
Hangzhou, 310058, China\\
\And
{Liang Xu}
\\
Zhejiang University of Technology\\
Hangzhou, 310014, China\\
\And
{Hailin Feng}
\\
Zhejiang Agriculture and Forestry University\\
Hangzhou, 311300, China\\
\And
{Zhenhong Du}
\\
Zhejiang University\\
Hangzhou, 310058, China\\
}

\renewcommand{\shorttitle}{\textit{ControlCity}}

\hypersetup{
pdftitle={A template for the arxiv style},
pdfsubject={q-bio.NC, q-bio.QM},
pdfauthor={David S.~Hippocampus, Elias D.~Striatum},
pdfkeywords={First keyword, Second keyword, More},
}

\begin{document}
\maketitle

\begingroup
\renewcommand{\thefootnote}{}
\footnotetext{This is a preprint of an article published by Taylor \& Francis in International Journal of Geographical Information Science (Received 2 Sep 2025; Accepted 25 Feb 2026), available at: 
https://doi.org/10.1080/13658816.2026.2639026}
\endgroup

\begin{abstract}
	Urban morphology is fundamental to determining urban functionality and vitality. Prevailing simulation methods, however, often oversimplify morphological generation as a geometric problem, lacking a profound understanding of urban semantics and geographical context. To address this limitation, this study proposes ControlCity, a diffusion model that achieves comprehensive urban morphology generation through multimodal information fusion. We first constructed a quadruple dataset comprising ``image-text-metadata-building footprints" from 22 cities worldwide. ControlCity utilizes these multidimensional information as joint control conditions, where an enhanced ControlNet architecture encodes spatial constraints from images, while text and metadata provide semantic guidance and geographical priors respectively, collectively directing the generation process. Experimental results demonstrate that compared to unimodal baselines, this method achieves significant advantages in morphological fidelity, with visual error (FID) reduced by 71.01\%, reaching 50.94, and spatial overlap (MIoU) improved by 38.46\%, reaching 0.36. Furthermore, the model demonstrates robust knowledge generalization and controllability, enabling cross-city style transfer and zero-shot generation for unknown cities. Ablation studies further reveal the distinct roles of images, text, and metadata in the generation process. This study confirms that multimodal fusion is crucial for achieving the transition from ``geometric mimicry" to ``understanding-based comprehensive generation," providing a novel paradigm for urban morphology research and applications.
\end{abstract}

\keywords{Diffusion Model \and Multimodal Fusion \and Controllable Generation \and Urban Morphology \and Geospatial AI}

\section{Introduction}

Urban morphology—the physical structure composed of streets, blocks, and buildings within cities—constitutes a core element that shapes human socioeconomic activities \citep{1-01, 1-02, 1-03}, influences residents' quality of life \citep{1-04, 1-05, 1-06}, and determines urban sustainability capacity \citep{1-07, 1-08, 1-09}. Accurate understanding, simulation, and generation of urban forms with specific characteristics are of paramount importance for fields such as urban planning, architectural design, and geographical information science in addressing challenges brought by rapid urbanization \citep{1-10, 1-11, 1-12}.

In recent years, the rapid advancement of generative artificial intelligence, particularly deep learning approaches, has provided unprecedented powerful tools for computational urban morphology research. Simultaneously, Volunteered Geographic Information (VGI) platforms, exemplified by OpenStreetMap (OSM), have provided these data-driven models with massive, open datasets containing rich multimodal information (geometric, semantic, attributional, etc.), significantly advancing the field of urban form generation \citep{1-13, 1-14, 1-15, 1-16, 1-17}.

However, despite continuous technological progress, existing methods tend to reduce urban form generation—essentially a ``complex systems synthesis problem"—to a ``geometric shape generation problem." This reductionist paradigm results in systematic neglect of two critical dimensions constituting urban morphology: semantics and context. This ``paradigmatic gap" manifests across various existing methodological approaches. Traditional procedural modeling approaches \citep{1-18pmcity, 1-19, 1-20} rely on manually defined hard-coded rules and are essentially pure geometric operations, completely lacking the capability to learn semantics and context from real-world data.

Emerging AI generation methods, exemplified by GANmapper \citep{1-21ganmapper} and InstantCITY \citep{1-22instantcity}, while achieving data-driven learning, nevertheless fall into two major predicaments—semantic void and contextual neglect—due to their reliance on unimodal information (e.g., road network images only). Since road network images inherently lack functional zoning information, these models cannot distinguish between ``commercial areas" and ``residential areas," rendering their generation process ``semantically blind." Due to the absence of macroscale geographical location awareness, models sever intrinsic connections between generation targets and broader geographical regions, producing ``context-insulated" results that fail to reflect historical and cultural differences across different regions. Even extensions to InstantCITY attempting to incorporate population and building density information are constrained by their paradigmatic limitations \citep{1-23}, requiring separate model training for different density categories, thereby increasing application complexity without fundamentally resolving the underlying issues. Ultimately, these approaches constitute pixel-based ``correlation mimicry" rather than multidimensional information-based ``systematic understanding generation."

In this paper, we propose ControlCity, a multimodal urban morphology synthesis model designed to address existing paradigmatic gaps. Our approach begins with constructing a multimodal aligned dataset capable of supporting deep morphological understanding. Through an innovative data processing pipeline, we extract and align four types of critical information from sources including OSM and Wikipedia for each geographical block across 22 global cities: road networks and land use images defining physical constraints, text injecting functional and stylistic information, metadata (coordinates) providing geographical context, and building footprints serving as generation targets. Building upon this foundation, the ControlCity model is designed to synergistically integrate these multidimensional information sources: during generation, an enhanced ControlNet encodes image modalities as rigid spatial ``skeletons"; a robust text encoder (CLIP) parses textual prompts into rich semantic instructions that guide the generation of building layouts conforming to specific functions and styles; finally, metadata provides macroscale geographical location priors through sinusoidal embeddings, ensuring generated results align with their regional ``context." Through deep fusion of these three components during the denoising process of the diffusion model, ControlCity generates precise and highly realistic building footprint data. 

A series of systematic experiments validate the effectiveness and superiority of our approach. First, through direct comparison with unimodal baselines, we confirm that multimodal information is crucial for enhancing morphological fidelity. Our method not only reduces error by 71.01\% in visual fidelity (average FID of 50.94) but also achieves a significant 38.46\% improvement in MIoU measuring spatial overlap (average MIoU of 0.36), while demonstrating superior performance in key morphological metrics, such as substantially reducing the absolute average error of $\mathit{\Delta} \text{ Site Cover}$ from 8.51\% to 3.82\%. Second, our model demonstrates remarkable flexibility in learned knowledge application. Results indicate that the model can successfully transfer and superimpose learned urban morphological styles onto novel geographical spatial structures according to textual instructions; simultaneously, when confronting completely unknown cities, it can robustly perform zero-shot generation by synthesizing its knowledge base, demonstrating powerful generalization capabilities. Finally, ablation studies precisely dissect the synergistic mechanisms of each modality, revealing the unique roles of image modalities as ``physical skeletons," textual prompts as ``semantic" cores, and geographical coordinates as ``global contextual calibration" for zero-shot generalization. 

In summary, the principal contributions of this study are as follows:
\begin{enumerate}
  \item Proposed a novel urban morphology comprehensive generation paradigm that achieves the transition from ``geometric mimicry" to ``understanding-based generation" through multimodal information fusion.
  \item Designed and validated a multimodal diffusion model named ControlCity that significantly outperforms existing unimodal methods in morphological fidelity, generalization, and controllability.
  \item Established a comprehensive multimodal geographical dataset processing and generation workflow that provides methodological references for future urban morphology research based on multidimensional information.
\end{enumerate}

\section{Related Work}

\subsection{Diffusion Model}

In recent years, diffusion models \citep{2-01} have achieved remarkable progress in image generation, surpassing previously dominant generative adversarial networks (GANs) \citep{2-02}, variational autoencoders (VAEs) \citep{2-03}, and flow-based models \citep{2-04}. Diffusion models transform Gaussian noise into images following target distributions through iterative denoising processes, encompassing both diffusion and denoising phases. Denoising Diffusion Probabilistic Models (DDPM) \citep{2-05ddpm} improved diffusion model training methodologies by utilizing variational inference to train parameterized Markov chains, thereby generating high-quality samples. To enhance sampling efficiency in diffusion models, Denoising Diffusion Implicit Models (DDIM) \citep{2-06ddim} introduced non-Markovian diffusion processes, substantially reducing required sampling steps and accelerating generation speed. \citet{2-07} further enhanced the flexibility and efficiency of generative models by introducing stochastic differential equations. Conditional diffusion models extend DDPM architectures, analogous to conditional GANs (cGANs) \citep{2-08} and conditional VAEs (cVAEs) \citep{2-09}, by modulating outputs based on additional input information.

In the domain of text-driven image generation, diffusion model-based approaches are currently considered the most promising. These approaches typically encode textual inputs into latent vectors through pretrained language models such as CLIP \citep{clip2-10}. GLIDE \citep{GLIDE2-11} employs cascaded diffusion architectures to achieve text-guided image generation and editing. Imagen \citep{Imagen2-12} leverages the powerful text comprehension capabilities of the large pretrained language model T5 \citep{2-13}, significantly improving high-fidelity image generation performance. The currently most popular text-to-image generation model is the Latent Diffusion Model (LDM) \citep{ldm2-14} implementation (a.k.a. Stable Diffusion), which is trained on the large-scale image and text dataset LAION-5B \citep{Laion-5b2-15}, performs diffusion processes in latent space, and introduces cross-attention-based control mechanisms to enhance traditional diffusion model capabilities. This approach has inspired a series of subsequent research aimed at improving text-to-image synthesis \citep{sdxl2-16, pixart2-17}. In this study, we further extend the powerful generative capabilities of SDXL to adapt to urban building morphology generation tasks.

\subsection{Controllable Image Synthesis}

Text-guided diffusion models have demonstrated certain capabilities in generating images that align with user expectations; however, they still lack fine-grained control capabilities in domain-specific tasks. Typically, such fine-grained control signals are expressed in image form, for example, using segmentation maps as control signals to determine the layout and shape of generated images \citep{2-18, 2-19, 2-20}, or utilizing sketches as structural information to precisely control image generation \citep{2-21, 2-22, 2-23}. Additionally, content control can be achieved by extracting semantic information from input images to generate images with personalized characteristics \citep{2-24, 2-25}. However, early control methods were often designed for specific tasks, and such task-specific designs limited their applicability within broader research communities. How to design a universal framework capable of supporting large-scale user adoption based on existing pretrained diffusion models (e.g., Stable Diffusion) became an urgent problem requiring resolution.

To meet practical demands, researchers rapidly initiated exploration of universal frameworks for handling different types of spatial conditions. T2I-Adapter \citep{2-26} achieves finer control over the generation process by aligning external control signals with the internal knowledge of pretrained text-to-image (T2I) diffusion models. ControlNet \citep{controlnet2-27} proposes creating a trainable copy of the UNet encoder, encoding additional conditional signals into latent representations and injecting them into the backbone of T2I diffusion models through zero convolution. IP-Adapter \citep{ipa2-28} utilizes CLIP to extract global semantic representations of images and achieves content control through decoupled cross-attention mechanisms. InstantID \citep{instantid2-29} employs innovative IdentityNet and lightweight image adapters to achieve personalized facial transfer. Uni-ControlNet \citep{2-30} and Ctrl-X \citep{2-31} achieve flexible combinations of structural control and semantic appearance control through different architectural designs. However, despite these methods improving image generation control capabilities to some extent, they remain insufficient when handling complex spatial conditions, particularly struggling to generate harmonious, natural, and morphologically accurate building footprints under multimodal geospatial conditions.

\subsection{Urban Building Layout Generation}

In the fields of urban planning and architectural design, traditional procedural urban modeling initially generated city layouts through manually coded rules and constraints, effectively ensuring topological structural rationality \citep{1-18pmcity}. Nevertheless, this manual process required designers to manually define rules and design options, limiting design flexibility and efficiency \citep{1-19, 1-20}. In recent years, with the rapid development of artificial intelligence generation technologies, researchers have begun exploring the use of generative models to automatically generate urban building layouts that meet specific requirements. LayoutGAN \citep{layoutgan2-32} learns layout design rules and characteristics through generative adversarial networks to generate image layouts conforming to design standards, while LayoutVAE \citep{2-33} learns distributions from existing layout data, enabling generation of layouts similar to input data with diversity and variations. These layout generation models were initially primarily applied to document and graphic layouts \citep{2-34, 2-35} but have gradually expanded to other domains. In interior design, House-GAN \citep{2-36} proposed a graph-constrained generative adversarial network for automatically generating diverse and realistic house layouts matching input bubble diagrams, while Graph2Plan \citep{2-37} combines generative models with user interaction to generate floor plans meeting user requirements based on input layout diagrams and architectural boundaries. Additionally, several studies \citep{2-38, 2-39} have employed deep generative models for indoor scene synthesis.

As research focus shifted toward larger-scale urban building layout synthesis, GAN-based models have been progressively applied to generate building layouts for different cities, demonstrating GANs' unique adaptability in learning urban morphology \citep{2-40, 1-21ganmapper}. BlockPlanner \citep{Blockplanner2-41} achieved diverse and effective urban block generation by introducing vectorized dual-layer graph representations. ESGAN \citep{2-42} generates visually realistic and semantically reasonable urban layouts by combining deep generative methods with urban conditional encoding. The method proposed by \citet{2-43} can generate realistic urban layouts based on arbitrary road networks, aiming to address limitations of existing methods in handling arbitrarily shaped urban blocks and diverse architectural forms. InstantCITY, proposed by \citet{1-22instantcity}, can generate high-resolution building vector data from street networks and demonstrated its application potential in urban geography through experiments across 16 global cities. Furthermore, \citet{1-23} extended InstantCITY by incorporating constraints such as population and building density. However, existing methods typically rely on unimodal data (i.e., road networks) for urban layout generation, and facing challenges when scaling to large metropolitan areas. To address this limitation, we propose ControlCity. Our research is inspired by the core concept of Geographical Data Translation pioneered by studies such as GANmapper and InstantCITY (i.e., utilizing one geographic feature to generate another related geographic feature), yet adopts a novel approach in both core technology and fundamental paradigm. We employ diffusion models as the generative core and innovatively introduce a multimodal fusion mechanism, aiming to achieve a paradigm shift from ``geometric mimicry" to ``understanding-based generation," while consolidating morphological knowledge from multiple cities into a single model, thereby providing the potential for learning global metropolitan morphological patterns.

\section{Methodology}

\subsection{Overview}
\begin{figure*}[!t]
	\centering
	\includegraphics[width=\textwidth]{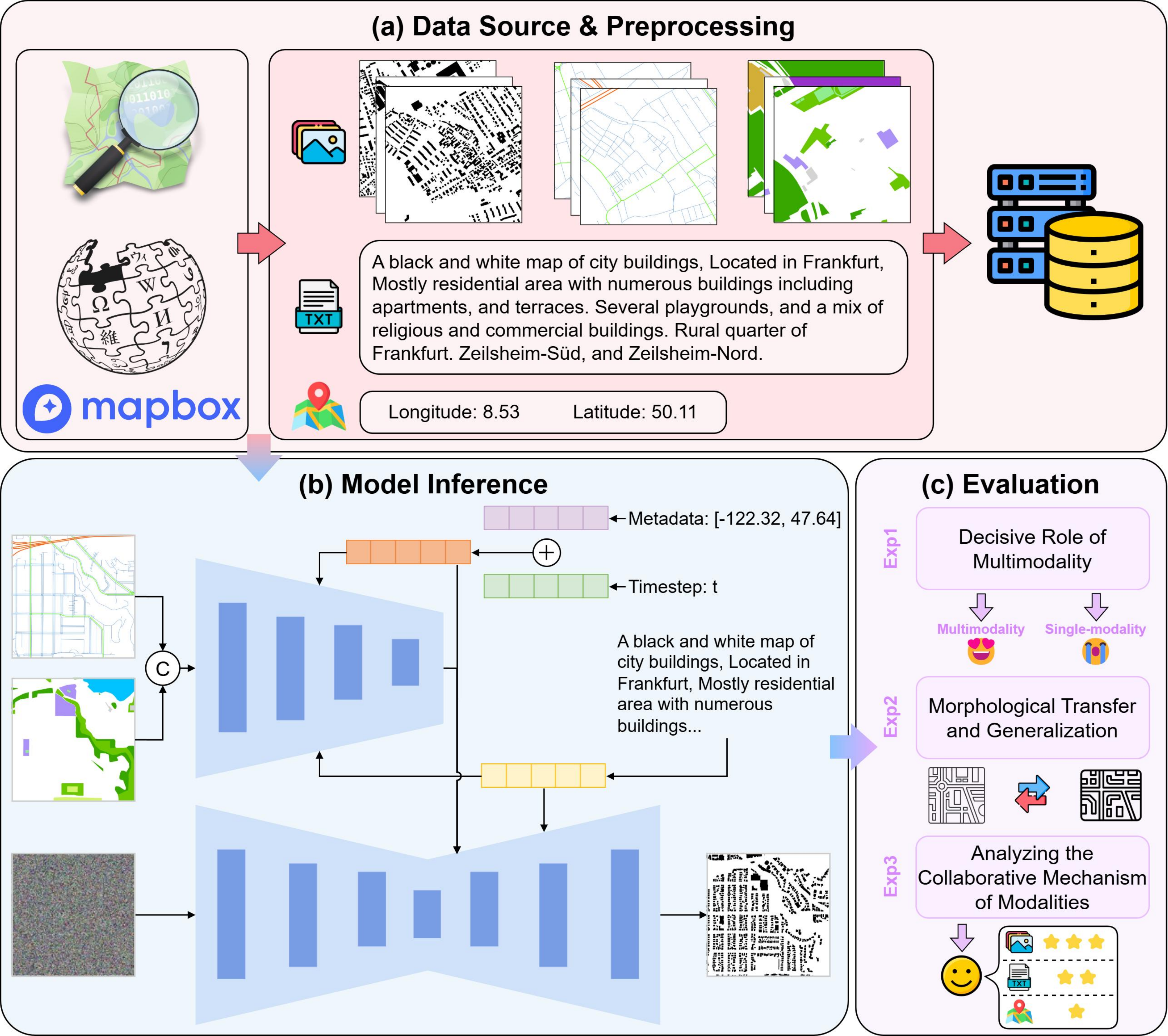}
	\caption{The overall framework of ControlCity, illustrating the three primary stages: (a) multimodal data preprocessing from geospatial sources, (b) the conditional diffusion model inference pipeline, and (c) the experimental evaluation process.}
	\label{fig1}
\end{figure*}

To achieve comprehensive understanding and generation of urban morphology, this study first proposes a multimodal alignment data construction pipeline. We processed data from 22 cities worldwide—including road networks, land use patterns, OSM attributes, Wikipedia content, and geographic coordinates for each tile—into images, text, and metadata formats, respectively. These were then aligned with target building footprints to construct a quadruple dataset comprising ``image-text-metadata-building footprint" that supports morphological comprehensive generation tasks.

Based on this multimodal dataset, we propose ControlCity, an enhanced pre-trained text-to-image generation model. The core mechanism of this model lies in its ability to synergistically fuse information from different modal sources: the ``physical skeleton" that defines spatial constraints (i.e., road networks and land use images), the ``semantic essence" that injects local style and functionality (i.e., textual prompts), and the ``geographic context" that provides macroscopic background information (i.e., coordinate data). Through this integration, the model generates urban morphologies with high fidelity and geographic authenticity.

In the experimental section, we validate the effectiveness of our approach through a series of rigorous tests. First, through direct comparison with unimodal baselines, we demonstrate the decisive role of multimodal information in enhancing morphological fidelity (Section~\ref{4-1}). Subsequently, we examine the generalization capability and controllability of the morphological knowledge learned by the model through style transfer and zero-shot generation tasks (Section~\ref{4-2}). Furthermore, we conduct ablation studies to precisely analyze the specific contributions of each modal information source (Section~\ref{4-3}). The overall framework is illustrated in Figure~\ref{fig1}.

\subsection{Dataset Construction}
\begin{figure*}[!t]
	\centering
	\includegraphics[width=\textwidth]{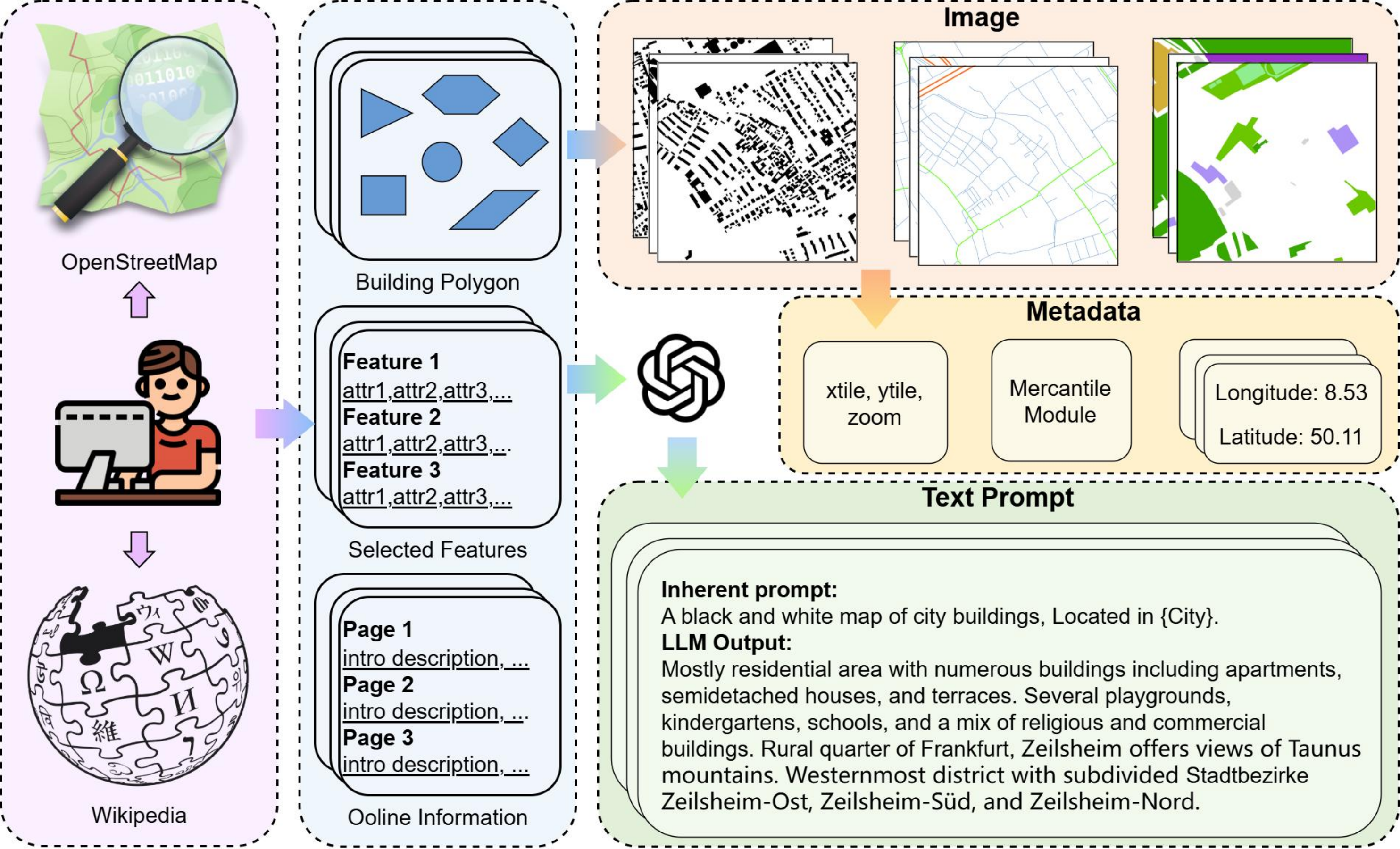}
	\caption{The multimodal data construction pipeline. Data from OpenStreetMap and Wikipedia are processed and fused using an LLM to generate an aligned dataset of images, text prompts, and metadata for each building footprint target.}
	\label{fig2}
\end{figure*}

The data foundation of this study is OSM, an open, free, and editable mapping project created and maintained by volunteers worldwide. Through multi-source data integration and community validation, OSM has established a detailed and accurate global geographic database that provides valuable resources for data-driven urban morphology research.

To train and evaluate a model with comprehensive morphological knowledge, we extracted representative urban data from 22 cities across different continents and countries from OSM. These cities are primarily economically developed regions with high building data completeness, whose diverse morphologies provide sufficient data support for model learning.

To capture macroscopic urban texture while preserving sufficient local detail, we adopted tiles at zoom level 15. We established a preprocessing pipeline to convert study areas into 1024×1024 pixel raster tiles, stored in Web Map Tile Service (WMTS) XYZ tile directories. Concurrently, we obtained corresponding street network maps and land use maps for each tile from MapBox. These two components collectively constitute the ``physical canvas" for morphology generation, defining the physical structure and functional zoning of building layouts.

However, the physical canvas alone is insufficient for comprehensive morphology generation. Therefore, we extracted rich semantic and contextual information from two complementary sources: OSM internal attributes and external knowledge bases (Wikipedia). In OSM, each geographic feature contains abundant semantic attributes (e.g., ``highway", ``landuse"). Based on relevance assessment, we selected 185 attributes related to building morphology and aggregated attributes within each tile through a preprocessing pipeline to form structured OSM-Caption features. To compensate for OSM's limitations in describing historical and cultural contexts, we utilized Wikipedia's GeoSearch functionality. By querying geographic entries within a 500-meter radius of each tile's center coordinates, we obtained more comprehensive community descriptions, forming Wikipedia-Caption features. Additionally, we calculated the center geographic coordinates of each XYZ tile as metadata to provide essential geographic context for the model.

The raw text directly extracted from OSM tags and Wikipedia descriptions suffers from information redundancy and structural inconsistency, making it unsuitable for direct model utilization. Existing research demonstrates \citep{pixart2-17, 3-01} that recaptioning using large language models (LLMs) is an effective information refinement method. Therefore, we employed GPT-4o mini to fuse and recaption the aforementioned two caption features, transforming them into concise, coherent, and information-dense textual prompts. Ultimately, we constructed a quadruple data processing pipeline comprising ``image-text-metadata-building footprint" (Figure~\ref{fig2}), which generated an aligned multimodal dataset containing 3,140 high-quality sample pairs, establishing a solid foundation for subsequent morphological synthesis model training.

\subsection{Model Architecture}
\begin{figure*}[!h]
	\centering
	\includegraphics[width=\textwidth]{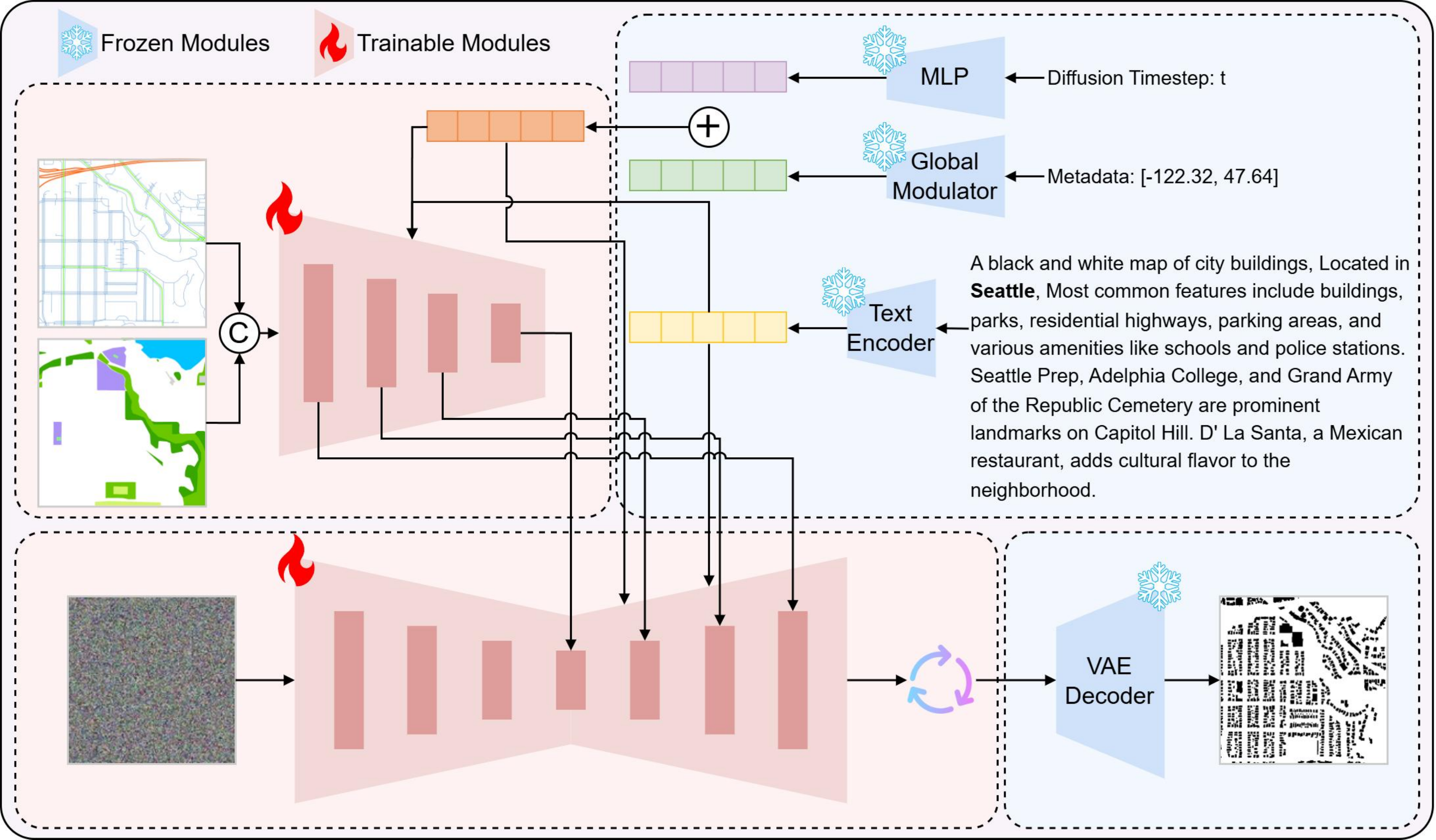}
	\caption{The ControlCity model architecture. A ControlNet injects spatial conditions (road and land use images) into a frozen SDXL UNet backbone. Text and metadata provide semantic and geographic guidance to control the generation process.}
	\label{fig3}
\end{figure*}

To achieve comprehensive generation of urban morphology, we propose ControlCity, a generative architecture deeply customized based on the advanced text-to-image diffusion model Stable Diffusion XL (SDXL). SDXL was selected as the foundation due to its powerful image generation capabilities and flexible multi-conditional control interfaces, providing an ideal platform for integrating different modal information. The core design philosophy of ControlCity is to synergistically inject different types of modal information into the generation process, collectively guiding the model to sample high-fidelity urban morphologies from noise.

The SDXL architecture comprises three core components: a Variational Autoencoder (VAE) for efficient conversion between pixel space and latent space; a powerful text encoder combination (OpenCLIP ViT-bigG and CLIP ViT-L) for understanding text prompts; and a UNet-based noise prediction network that performs iterative denoising in the model's latent space, serving as the core of the generation process.

To efficiently encode the ``physical skeleton" of morphology generation—i.e., road networks and land use maps—as spatial constraints, we improved the existing ControlNet architecture. The traditional approach involves training independent ControlNets for street network maps $c_s$ and land use maps $c_l$ separately. However, this method struggles to capture the intrinsic relationships between these two types of spatial information. Our improvement lies in first concatenating the street network and land use maps along the channel dimension to form a composite image condition $c_i$:$c_i=Concat(c_s,c_l)$. Subsequently, this is jointly encoded into feature maps $c_f$ through a lightweight convolutional network: $c_f$:$c_f=Conv(c_i)$. These feature maps extract multi-scale features of the input image through three downsampling blocks and one middle block $F_c = \{ F_{Down}, F_{CADown}, F_{CADown}, F_{Mid} \}$ and are injected at multiple levels of the UNet through zero convolutions to guide the denoising process. Compared to independent encoding, this joint encoding strategy enables the model to learn the synergistic relationship between road network structure and land use zoning at an early stage, thereby generating more coordinated and natural building layouts.

Beyond the aforementioned image-based spatial constraints, the model also integrates two types of non-visual conditions—text and metadata—to provide richer generative guidance. Text semantic information, contained in prompts rich with urban information, is encoded through pre-trained text encoders and guides generation during the UNet denoising process via cross-attention layers, endowing the generated results with specific stylistic and functional characteristics. For geographic coordinates, to avoid the inherent limitations of text encoders in processing numerical information, we employed the same sinusoidal embedding used for timestep embedding \citep{transformer, diffusionsat}. The specific formula is as follows:

\begin{equation}
    \text{Proj}(k, 2i) = \sin\!\left(m \Omega^{- \tfrac{2\pi}{d}} \right), \quad
    \text{Proj}(m, 2i+1) = \cos\!\left(k \Omega^{- \tfrac{2\pi}{d}} \right)
\end{equation}

\begin{equation}
    \mathbf{c}_{m,t} = \sum_{j=1}^{2} \text{MLP}\left( [\text{Proj}(m_j, 0), \dots, \text{Proj}(m_j, d)] \right) + \text{MLP}\left( [\text{Proj}(t, 0), \dots, \text{Proj}(t, d)] \right)
\end{equation}
where $i$ is the index of the feature dimension and $\Omega=1000$. Each coordinate value is projected through a Global Contextual Modulator and then added to the timestep embedding to generate the final conditional vector $c_{m,t}$, providing the model with macroscopic geographic location priors.

At each denoising step, the spatial structural constraints injected by the improved ControlNet, the semantic style instructions injected through cross-attention layers, and the geographic location priors embedded together with timesteps undergo deep fusion across various levels of the UNet. The model's objective is to predict and remove noise, thereby progressively recovering the target image $x$ representation $z_{t_0}^{(x)}$  in latent space from Gaussian noise under the collective guidance of all modal conditions. Finally, the VAE decoder $\mathcal{D}$ reconstructs this latent representation into high-resolution pixel images $\hat{x} = \mathcal{D}(z_{t_0}^{(x)})$. The overall model architecture is illustrated in Figure~\ref{fig3}.

\subsection{Evaluation Metrics}\label{3-4}

To comprehensively and objectively evaluate the performance of our model, we established an evaluation framework encompassing two dimensions: visual quality and urban morphology. This framework aims to ensure that generated images not only exhibit high visual fidelity but also accurately reflect real-world urban building morphology.

For visual quality metrics, we employed the Fréchet Inception Distance (FID) \citep{fid} as a standard benchmark for evaluating the overall visual quality of generated images. FID measures image quality by calculating the distributional differences between generated and real images in the feature space of a pre-trained InceptionV3 network. A lower FID score indicates greater similarity between the two image distributions, thereby signifying higher visual quality of the generated images. The calculation formula is as follows:
\begin{equation}
    \text{FID} = \| \mu_r - \mu_g \|_2^2 + {Tr}(\Sigma_r + \Sigma_g - 2(\Sigma_r \Sigma_g)^{\tfrac{1}{2}})
\end{equation}
where $\mu_r$ and $\mu_g$ represent the mean vectors of real and generated image features, respectively, and $\Sigma_r$ and $\Sigma_g$ are the corresponding covariance matrices.

Beyond macro-level visual quality, we focus more on the accuracy of generated results at the micro-level of building morphology. To this end, we vectorized the generated raster images into geospatial polygons to conduct a series of quantitative morphological assessments at the GIS level. These metrics are widely utilized in urban morphology research \citep{1-22instantcity, metric2}. Mean Intersection over Union (MIoU) measures the overlap between generated and real buildings at the raster level. A higher MIoU score (with an ideal value of 1) indicates that generated building outlines more closely resemble real buildings in terms of shape and position. $\mathit{\Delta}$ Building Area and $\mathit{\Delta}$ Building Perimeter measure the average morphology of individual buildings. These metrics calculate the percentage differences between the average area/perimeter of generated buildings and real buildings within each tile, respectively. The ideal value for both metrics is 0\%, indicating that generated individual buildings are similar to real buildings in terms of average size and shape complexity. $\mathit{\Delta} \text{ Site Cover}$ measures the percentage difference between the total generated building area and total real building area relative to the total tile area within each tile. The ideal value is 0\%, indicating that the total volume of generated buildings matches the real situation. GN Count measures the ratio between the number of generated building entities and the actual number. The ideal value is 100\%, indicating that the model accurately generates the same number of buildings as in reality.
\begin{equation}
\text{MIoU} = \frac{\text{Bldg}_{GN} \cap \text{Bldg}_{GT}}{\text{Bldg}_{GN} \cup \text{Bldg}_{GT}}
\end{equation}
\begin{equation}
\mathit{\Delta}\ \text{Bldg. Area} = 100 \times \left( \frac{ \frac{\text{Bldg Area}_{GN}}{\text{Bldg Count}_{GN}} }{ \frac{\text{Bldg Area}_{GT}}{\text{Bldg Count}_{GT}} } - 1 \right)
\end{equation}
\begin{equation}
\mathit{\Delta}\ \text{Bldg. PM} = 100 \times \left( \frac{ \frac{\text{Bldg Perimeter}_{GN}}{\text{Bldg Count}_{GN}} }{ \frac{\text{Bldg Perimeter}_{GT}}{\text{Bldg Count}_{GT}} } - 1 \right)
\end{equation}
\begin{equation}
\mathit{\Delta}\ \text{Site Cover} = 100 \times \left( \frac{\text{Bldg Area}_{GN} - \text{Bldg Area}_{GT}}{\text{Tile Area}} \right)
\end{equation}
\begin{equation}
\text{GN Count} = 100 \times \frac{\text{Bldg Count}_{GN}}{\text{Bldg Count}_{GT}}
\end{equation}

\subsection{Experiment Setup}

To systematically validate the effectiveness of our proposed multimodal comprehensive generation method and its underlying mechanisms, we designed a series of sequential experiments. We first validate the decisive role of multimodal information in Experiment 1 through direct comparison between our multimodal model and a unimodal baseline across 10 global cities, aiming to demonstrate that multimodal information fusion is a critical factor for achieving high-fidelity morphological generation. Building upon this foundation, Experiment 2 further examines the generalization and controllability of the morphological knowledge learned by the model. This experiment evaluates the model's ability to flexibly apply its knowledge base for reasonable morphological inference through two tasks: morphological style transfer and zero-shot city generation. Finally, Experiment 3 conducts an ablation study to deeply analyze the specific contributions of each modality, aiming to open the model's ``black box" and mechanistically understand the underlying reasons for our method's success.

\section{Experiment and Results}

\subsection{Experiment 1 - Evaluating Multimodal Generation Against a Unimodal Baseline}\label{4-1}
\begin{figure*}[!h]
	\centering
	\includegraphics[width=\textwidth]{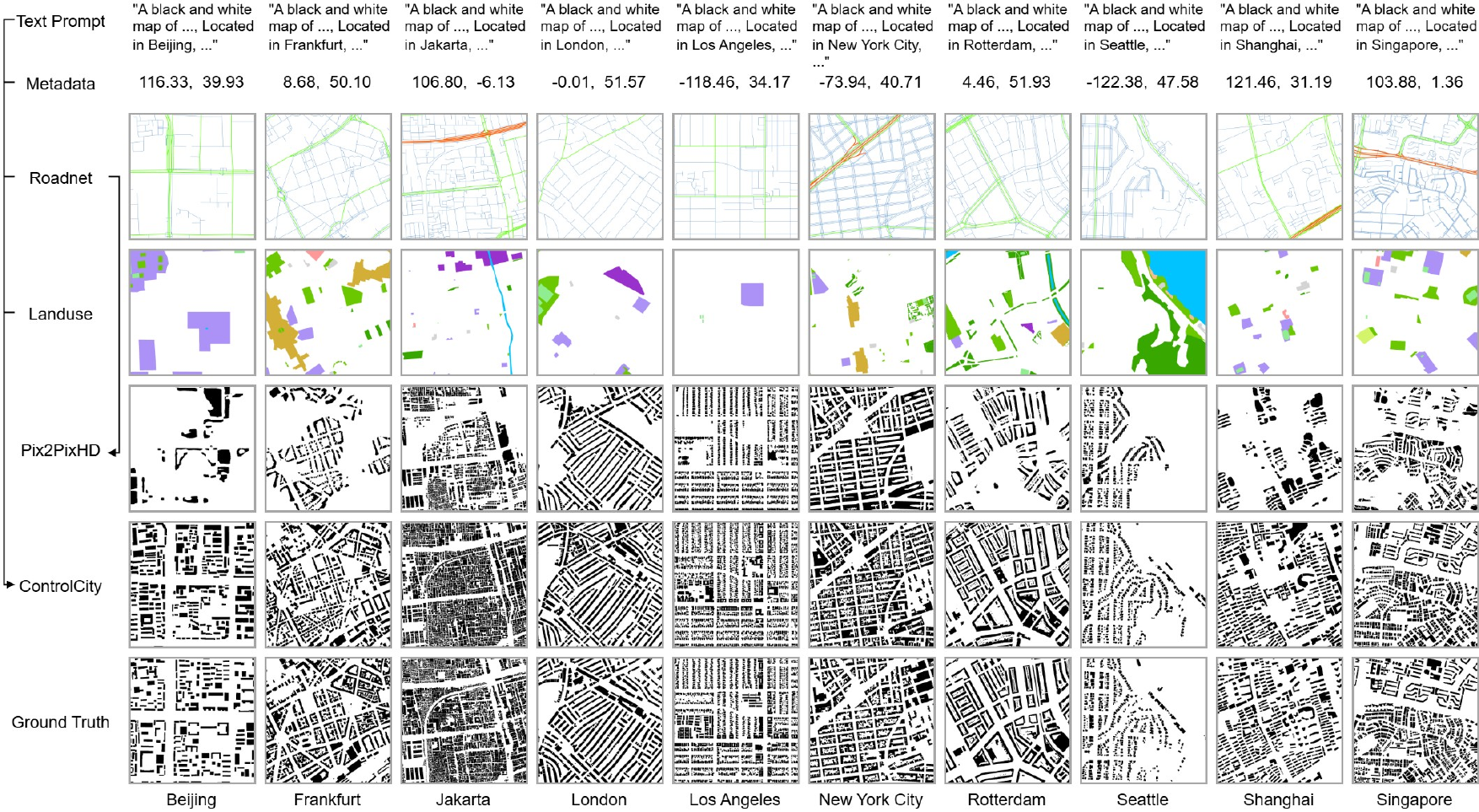}
	\caption{Qualitative comparison of generated building footprints across ten global cities. The results from ControlCity demonstrate significant improvements in morphological fidelity over the unimodal baseline (Pix2PixHD) when compared to the ground truth. The baseline was conditioned only on roadnet, while ControlCity used all multimodal inputs (roadnet, landuse, text prompt, and metadata). Detailed text prompts are provided in Appendix~\ref{appendix-a} (Table~\ref{tbl5}) .}
	\label{fig4}
\end{figure*}

This experiment aims to validate the core scientific hypothesis of our research: compared to generative modeling approaches that rely on a single visual modality, the comprehensive generation paradigm that integrates multidimensional information serves as the key driving force for achieving high-fidelity urban morphological generation. To this end, we constructed a controlled experiment. We selected 10 morphologically diverse global cities (i.e., Beijing, Frankfurt, Jakarta, London, Los Angeles, New York City, Rotterdam, Seattle, Shanghai, and Singapore), with all data sourced from OSM and processed into raster tiles, each covering approximately 1200m×1200m.

To evaluate the performance of our multimodal approach against current mainstream techniques, we selected Pix2PixHD \citep{wang2018pix2pixHD} as the unimodal generation baseline. This model represents the core generative methodology employed by advanced urban morphology generation studies such as InstantCITY \citep{1-22instantcity}, namely cGAN-based image translation. By reproducing this baseline under controlled experimental conditions, we were able to conduct direct comparisons against the mainstream, purely image-driven generative paradigm. In contrast, the proposed ControlCity integrates road networks, land use patterns, textual semantics, and geographical metadata within a unified framework, aiming to achieve comprehensive urban morphology generation.

Figure~\ref{fig4} intuitively demonstrates the decisive role of multimodal information in enhancing morphological fidelity. As a unimodal baseline, Pix2PixHD can only generate basic external contours in cities with clear road network structures and simple building layouts (e.g., Los Angeles and New York City), but completely fails in areas with complex morphology or unclear road network constraints (e.g., Beijing and Shanghai), resulting in extensive building omissions. This exposes the vulnerability of paradigms that rely solely on single visual cues (road network structure) when confronting complex real-world scenarios.

In contrast, ControlCity precisely captures the morphological ``DNA" of different cities through multidimensional information fusion: from Beijing's regular grid layout to Frankfurt's irregular courtyard style, and Jakarta's dense clusters of small buildings. This is not coincidental but rather the result of synergistic multimodal conditioning. This clearly indicates that beyond the fundamental physical constraints of road networks, the additionally integrated multimodal information — namely textual prompts containing urban functions, neighborhood characteristics, and cultural contexts, land use maps carrying functional zoning information, and coordinate data defining geographical context — collectively provide the model with deeper morphological cues.

\begin{table}
\centering
\caption{Quantitative comparison of model performance across ten global cities. The metrics evaluate both overall visual quality and the accuracy of key urban morphology characteristics, comparing ControlCity against the Pix2PixHD baseline.}
{
\begin{tabular}{clcccccc}
\toprule
\textbf{Model} &
  \textbf{City} &
  \textbf{FID} &
  \textbf{MIoU} &
  \textbf{\begin{tabular}[c]{@{}c@{}}$\mathit \Delta$ Bldg. \\ Area (\%)\end{tabular}} &
  \textbf{\begin{tabular}[c]{@{}c@{}}$\mathit \Delta$ Bldg. \\ PM (\%)\end{tabular}} &
  \textbf{\begin{tabular}[c]{@{}c@{}}$\mathit \Delta$ Site \\ Cover (\%)\end{tabular}} &
  \textbf{\begin{tabular}[c]{@{}c@{}}GN \\ Count (\%)\end{tabular}} \\ \midrule
\multirow{10}{*}{ControlCity} & Beijing       & 55.13           & 0.19          & 20.35          & 15.21          & 8.90          & 142.12          \\
                              & Frankfurt     & 53.36           & 0.31          & -28.85         & -8.52          & -0.03         & 139.73          \\
                              & Jakarta       & 74.38           & 0.41          & -7.22          & -1.50          & 6.03          & 131.67          \\
                              & London        & 43.93           & 0.45          & -18.83         & 1.33           & -1.94         & 111.01          \\
                              & Los Angeles   & 28.56           & 0.41          & -41.44         & -12.56         & 1.15          & 140.55          \\
                              & New York City & 37.81           & 0.50          & -37.77         & -23.03         & -0.24         & 172.98          \\
                              & Rotterdam     & 76.24           & 0.38          & -48.27         & -24.24         & -3.56         & 173.40          \\
                              & Seattle       & 47.80           & 0.40          & -59.32         & -31.66         & -2.74         & 172.50          \\
                              & Shanghai      & 43.04           & 0.20          & 4.52           & 12.44          & 9.55          & 157.86          \\
                              & Singapore     & 49.11           & 0.37          & 3.47           & 6.25           & 4.00          & 110.25          \\ \midrule
\textbf{Average}              & \textbf{}     & \textbf{50.94}  & \textbf{0.36} & \textbf{27.00} & \textbf{13.67} & \textbf{3.81} & \textbf{145.21} \\ \midrule
\multirow{10}{*}{Pix2PixHD}   & Beijing       & 337.52          & 0.12          & 159.05         & 46.99          & -5.25         & 30.22           \\
                              & Frankfurt     & 245.10          & 0.19          & -16.70         & -19.93         & -12.50        & 66.01           \\
                              & Jakarta       & 84.46           & 0.34          & 81.75          & 13.48          & -3.88         & 74.68           \\
                              & London        & 272.89          & 0.14          & 58.00          & 4.90           & -23.09        & 21.66           \\
                              & Los Angeles   & 61.35           & 0.31          & -25.84         & -12.71         & -9.13         & 83.74           \\
                              & New York City & 80.17           & 0.51          & -8.59          & -12.86         & -2.90         & 140.21          \\
                              & Rotterdam     & 217.29          & 0.23          & -43.75         & -41.32         & -11.88        & 120.09          \\
                              & Seattle       & 87.80           & 0.34          & -31.68         & -19.77         & -6.53         & 104.17          \\
                              & Shanghai      & 218.76          & 0.14          & 16.15          & 21.55          & -5.21         & 70.86           \\
                              & Singapore     & 151.89          & 0.28          & 12.15          & 5.83           & -4.70         & 70.51           \\ \midrule
\textbf{Average}              & \textbf{}     & \textbf{175.72} & \textbf{0.26} & \textbf{45.37} & \textbf{19.93} & \textbf{8.51} & \textbf{78.22}  \\ \bottomrule
\end{tabular}
}
\label{tbl1}
\end{table}

To quantitatively measure the performance gains brought by multimodal information, we employed the series of visual and urban morphological metrics defined in Section~\ref{3-4} for evaluation, with results presented in Table~\ref{tbl1}. When calculating averages, $\mathit{\Delta}$Bldg. Area, $\mathit{\Delta}$Bldg. PM, and $\mathit{\Delta} \text{ Site Cover}$ all use their absolute values for averaging to reflect overall error levels. All metrics indicate that the multimodal approach (ControlCity) significantly outperforms the unimodal baseline (Pix2PixHD) across all indicators. In terms of core visual fidelity, the multimodal approach achieved an average FID score of 50.94, compared to the unimodal baseline's 175.72, representing a substantial error reduction of 71.01\%. Regarding morphological metrics, ControlCity achieved average building area differences ($\mathit{\Delta}$Bldg. Area) and site coverage differences ($\mathit{\Delta} \text{ Site Cover}$) of 27.00\% and 3.81\%, respectively, substantially outperforming the unimodal baseline's 45.37\% and 8.51\%. This demonstrates that multimodal information can more effectively control the volume and density of generated buildings, making them closer to real conditions. For the MIoU metric measuring building distribution overlap, ControlCity achieved an average of 0.36, significantly higher than the unimodal baseline's 0.26. This order-of-magnitude improvement powerfully demonstrates the necessity of integrating multidimensional information for generating high-quality, visually realistic urban imagery.

\begin{figure*}[!t]
	\centering
	\includegraphics[width=\textwidth]{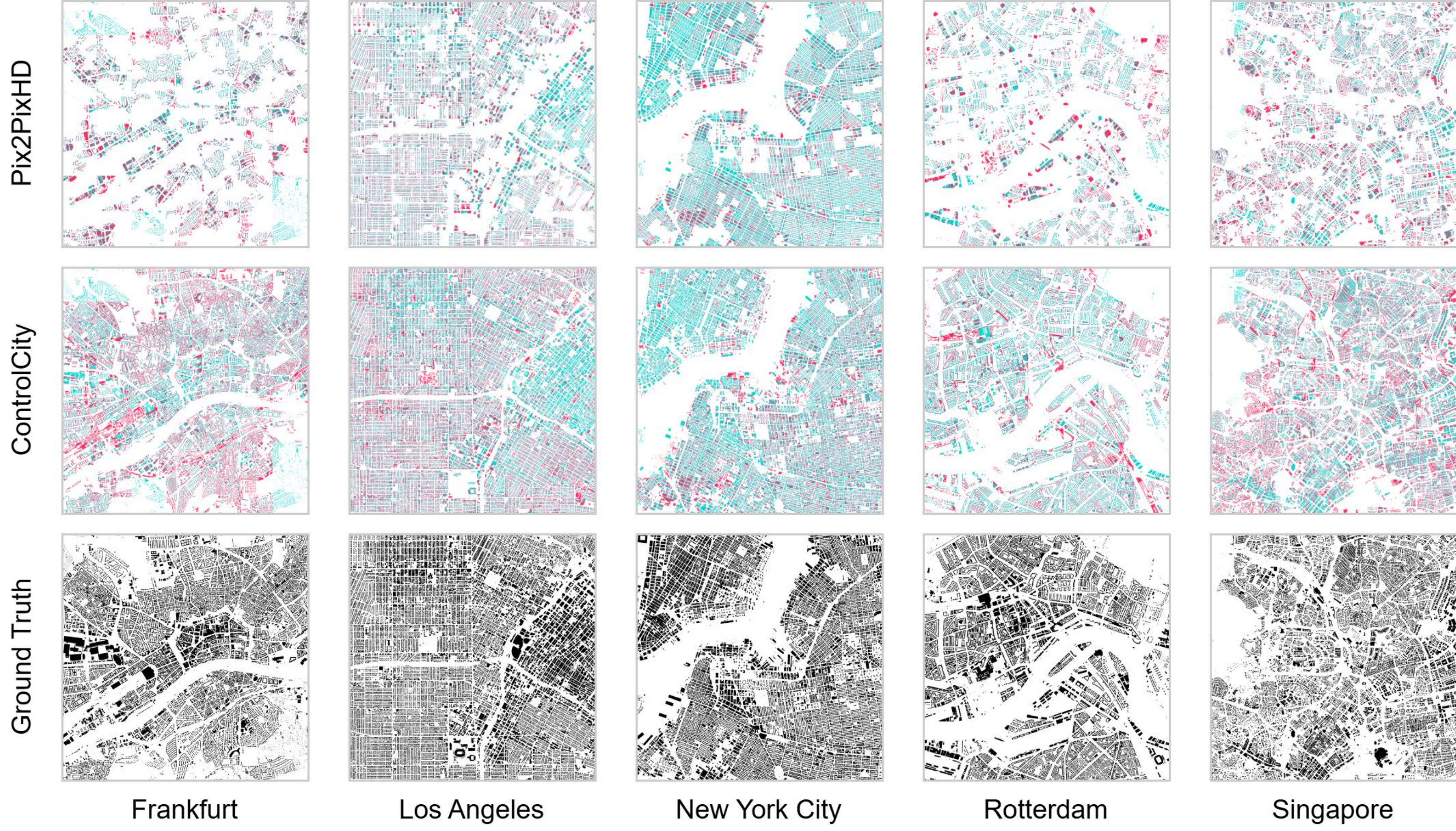}
	\caption{Large-scale generation results for five cities. Green indicates correctly generated buildings (True Positives) and red indicates errors (False Positives). The unimodal Pix2PixHD results show extensive blank areas (missed detections), while ControlCity demonstrates superior continuity and robustness.}
	\label{fig5}
\end{figure*}

To further evaluate the model's generation stability and continuity over large-scale areas, we assembled 7×7 tiles and visualized the generation results (Figure~\ref{fig5}). In the figure, green represents correctly generated buildings (true positives, TP), while red represents incorrectly generated buildings (false positives, FP). It can be clearly observed that the generation results of the unimodal method (Pix2PixHD) contain numerous scattered red noise points and extensive blank areas (missed detections), particularly in morphologically complex cities such as Frankfurt, where the results are fragmented and discontinuous. In contrast, ControlCity's generation results are more complete and continuous in morphology, with broader green area coverage and significantly reduced red error generation zones, demonstrating stronger robustness and generation consistency in complex urban environments.

In summary, whether through qualitative visual comparison, quantitative metric evaluation, or large-scale stability analysis, all evidence clearly points to the same conclusion: the integration of multimodal information is crucial for achieving high-fidelity urban morphological generation. This enables generative models to break free from dependence on single visual cues and understand and synthesize realistic, reasonable, and stable urban textures from deeper semantic and contextual levels, thereby validating the core hypothesis proposed in this research.

\subsection{Experiment 2 - Transfer and Generalization of Morphological Knowledge}\label{4-2}

In Experiment 4.1, we confirmed that multimodal information is crucial for achieving high-fidelity morphological generation. Building upon this foundation, this section further explores whether our model learns fixed, rigid transformation rules or higher-level, generalizable, and controllable ``morphological knowledge." We validate this through two sequential tests: first, examining whether the model can controllably transfer learned morphological styles to entirely new geographical canvases based on textual instructions; second, evaluating whether the model can generalize its aggregated knowledge base from multiple training cities when confronted with completely unknown cities.

\begin{figure*}[!t]
	\centering
	\includegraphics[width=\textwidth]{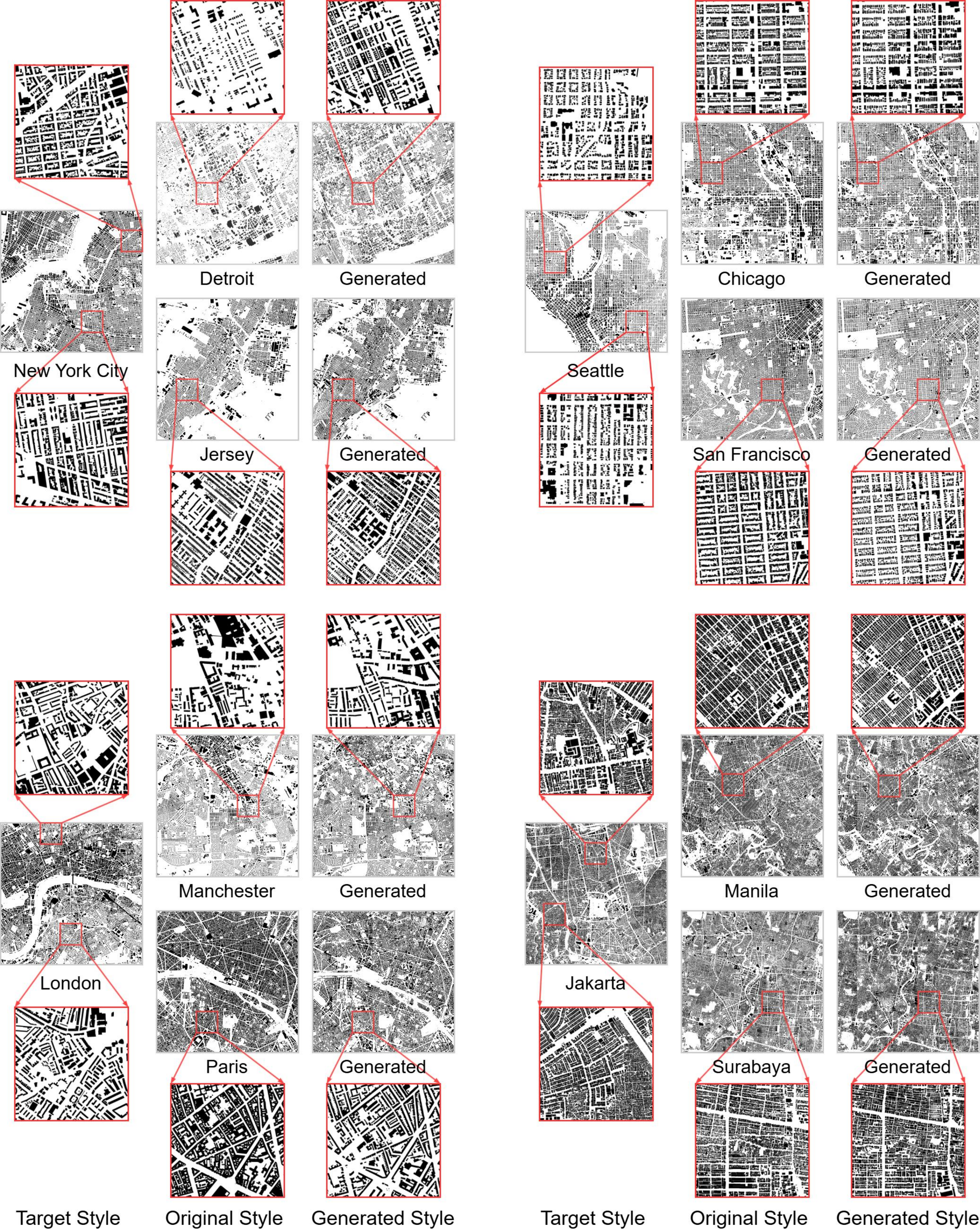}
	\caption{Morphological style transfer results. The model applies the style of a source city (e.g., London) to the geographical canvas of a target city (e.g., Paris), demonstrating effective stylistic control via text prompts.}
	\label{fig6}
\end{figure*}

In the first test regarding controllable transfer, we aim to evaluate the model's ability to successfully apply architectural morphologies learned from one city to other cities. We achieve this by fixing the textual prompt (e.g., using ``London" as a style instruction) while inputting road networks, land use, and coordinate information from an entirely new target city (such as Paris) to examine whether the model can transfer the source city's morphological style onto the target city's physical ``canvas." As shown in Figure~\ref{fig6}, the experimental results are compelling. In the ``London to Paris" case, the model did not simply replicate Paris's actual Haussmannian regular layout, but successfully superimposed London's distinctive, somewhat irregular courtyard-style architectural features onto Paris's grid-like road network. This clearly demonstrates that the textual modality plays a powerful ``style instruction" role in our model, achieving effective control over generation style.

\begin{table}[h]
\centering
\caption{Quantitative results for the morphological style transfer task, comparing the performance of ControlCity and Pix2PixHD.}
{
\begin{tabular}{cllcccccc}
\toprule
\textbf{Model}               & \textbf{City}                  & \textbf{Applied City} & \textbf{FID}    & \textbf{MIoU} & \textbf{\begin{tabular}[c]{@{}c@{}}$\mathit \Delta$ Bldg. \\ Area (\%)\end{tabular}} & \textbf{\begin{tabular}[c]{@{}c@{}}$\mathit \Delta$ Bldg. \\ PM (\%)\end{tabular}} & \textbf{\begin{tabular}[c]{@{}c@{}}$\mathit \Delta$ Site \\ Cover (\%)\end{tabular}} & \textbf{\begin{tabular}[c]{@{}c@{}}GN \\ Count (\%)\end{tabular}} \\ \midrule
\multirow{8}{*}{ControlCity} & \multirow{2}{*}{New York City} & Detroit               & 61.71           & 0.30          & 0.68                                                                & 24.97                                                             & 8.70                                                                & 149.91                                                           \\
                             &                                & Jersey                & 69.68           & 0.35          & -37.69                                                              & -12.35                                                            & 6.06                                                                & 233.95                                                           \\
                             & \multirow{2}{*}{Seattle}       & Chicago               & 42.67           & 0.44          & -45.54                                                              & -23.36                                                            & 0.18                                                                & 165.66                                                           \\
                             &                                & San Francisco         & 82.52           & 0.39          & -73.68                                                              & -58.57                                                            & -5.72                                                               & 275.67                                                           \\
                             & \multirow{2}{*}{London}        & Manchester            & 57.85           & 0.33          & -0.06                                                               & 20.26                                                             & 6.18                                                                & 133.22                                                           \\
                             &                                & Paris                 & 80.26           & 0.50          & -55.35                                                              & -21.33                                                            & -13.12                                                              & 169.64                                                           \\
                             & \multirow{2}{*}{Jakarta}       & Manila                & 64.02           & 0.39          & -16.43                                                              & -13.24                                                            & -0.49                                                               & 124.35                                                           \\
                             &                                & Surabaya              & 83.99           & 0.38          & 49.20                                                               & 26.60                                                             & 8.24                                                                & 92.81                                                            \\ \midrule
\textbf{Average}             & \textbf{}                      & \textbf{}             & \textbf{67.84}  & \textbf{0.39} & \textbf{34.83}                                                      & \textbf{25.09}                                                    & \textbf{6.09}                                                       & \textbf{168.15}                                                  \\ \midrule
\multirow{8}{*}{Pix2PixHD}   & \multirow{2}{*}{New York City} & Detroit               & 105.11          & 0.22          & 27.50                                                               & 32.02                                                             & 1.98                                                                & 90.92                                                            \\
                             &                                & Jersey                & 139.97          & 0.32          & -4.31                                                               & -1.43                                                             & -3.60                                                               & 120.30                                                           \\
                             & \multirow{2}{*}{Seattle}       & Chicago               & 66.61           & 0.33          & -37.89                                                              & -28.14                                                            & -7.93                                                               & 116.36                                                           \\
                             &                                & San Francisco         & 142.25          & 0.26          & -57.26                                                              & -53.00                                                            & -13.38                                                              & 160.22                                                           \\
                             & \multirow{2}{*}{London}        & Manchester            & 300.96          & 0.13          & 222.86                                                              & 64.59                                                             & -7.05                                                               & 23.94                                                            \\
                             &                                & Paris                 & 377.08          & 0.14          & -30.24                                                              & -32.81                                                            & -38.51                                                              & 34.29                                                            \\
                             & \multirow{2}{*}{Jakarta}       & Manila                & 123.27          & 0.32          & 67.62                                                               & 7.12                                                              & -10.19                                                              & 62.36                                                            \\
                             &                                & Surabaya              & 107.40          & 0.31          & 158.41                                                              & 49.67                                                             & -0.97                                                               & 47.99                                                            \\ \midrule
\textbf{Average}             & \textbf{}                      & \textbf{}             & \textbf{170.33} & \textbf{0.25} & \textbf{75.76}                                                      & \textbf{33.60}                                                    & \textbf{10.45}                                                      & \textbf{82.05}                                                   \\ \bottomrule
\end{tabular}
}
\label{tbl2}
\end{table}

The quantitative results in Table~\ref{tbl2} again introduce the unimodal baseline (Pix2PixHD) as comparison. The results show that the multimodal model (ControlCity) similarly demonstrates overwhelming advantages in style transfer tasks. Its average FID (67.84) is significantly lower than the unimodal baseline (170.33), indicating that its generated cross-style urban morphologies are visually more realistic and credible. The unimodal baseline almost completely fails when processing complex morphologies such as London and Paris, with FID scores reaching staggering levels of 300.96 and 377.08, respectively, and generated building counts (GN Count) of only 23.94\% and 34.29\% of real conditions, indicating complete model collapse.

In contrast, ControlCity maintained robust generation capabilities across all transfer tasks; for example, when transferring Seattle style to Chicago, it achieved an FID score as low as 42.67. More interestingly, in the ``London to Paris" case, its MIoU reached 0.50, the highest among all transfer tasks, indicating that the model effectively utilized precise spatial information from the target road network while superimposing styles.

\begin{figure*}[!t]
	\centering
	\includegraphics[width=\textwidth]{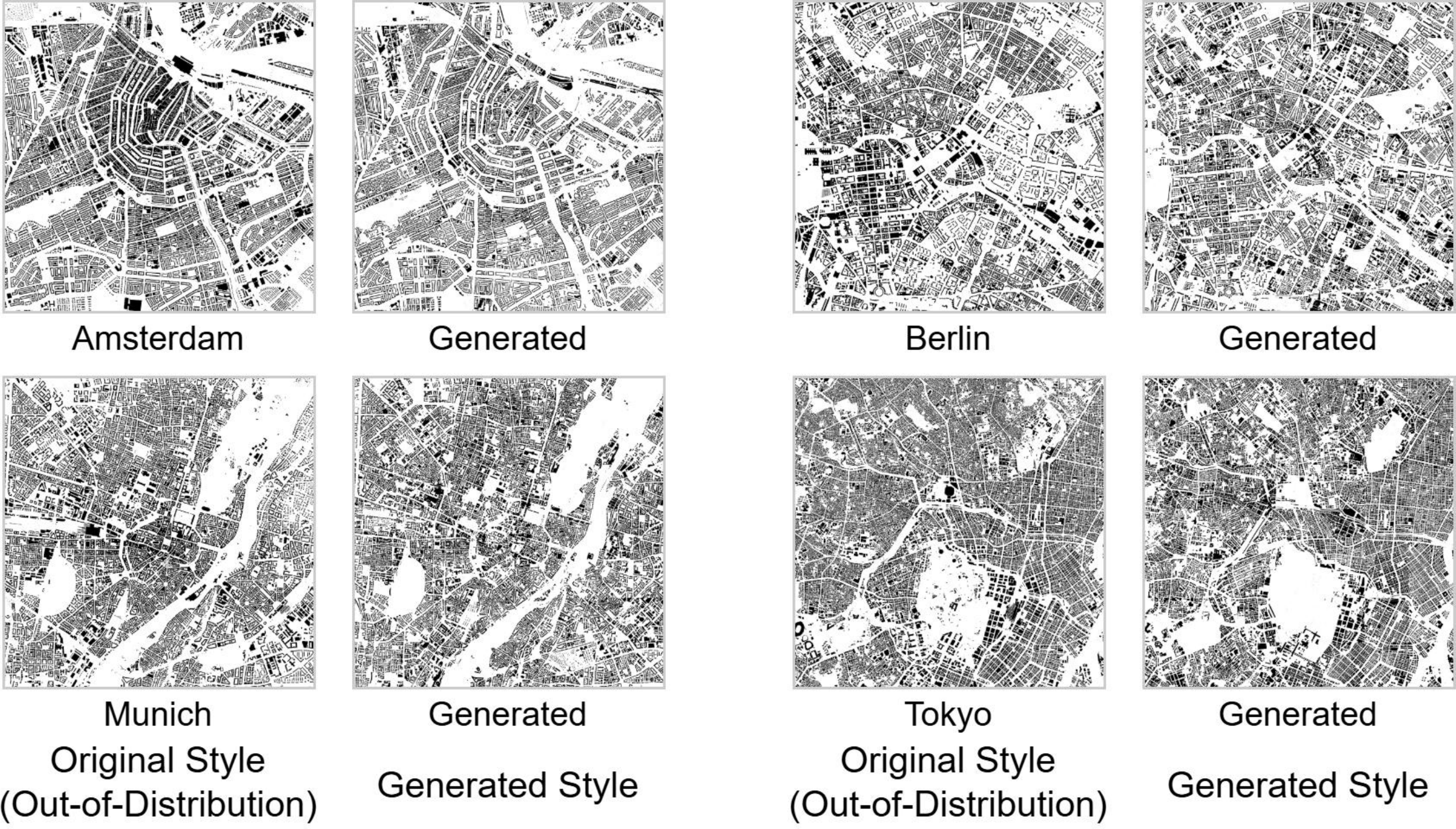}
	\caption{Zero-shot generation for four cities unseen during training. The model synthesizes morphologically plausible building layouts for out-of-distribution cities, showcasing its generalization capabilities.}
	\label{fig7}
\end{figure*}

Beyond controllable style transfer, we further tested the generalization limits of the model's knowledge base — namely, zero-shot generation capability. We selected four cities that never appeared during the training phase — Amsterdam, Berlin, Munich, and Tokyo — for testing, aiming to evaluate the model's ability to invoke its multimodal knowledge base learned from 22 different training cities for comprehensive inference. As shown in Figure~\ref{fig4}, although the model had never ``seen" these cities, it successfully generated morphologically highly reasonable building layouts for each new city. For example, the model accurately captured Amsterdam's canal-driven radial urban texture, as well as Berlin's and Munich's urban morphologies that combine regular blocks with irregular green spaces.

\begin{table}
\centering
\caption{Quantitative results for the zero-shot generation task on four out-of-distribution cities not included in the training set.}
{
  \begin{tabular}{lcccccc}
\toprule
\textbf{City} &
  \textbf{FID} &
  \textbf{MIoU} &
  \textbf{\begin{tabular}[c]{@{}c@{}}$\mathit \Delta$ Bldg. \\ Area(\%)\end{tabular}} &
  \textbf{\begin{tabular}[c]{@{}c@{}}$\mathit \Delta$ Bldg. \\ PM(\%)\end{tabular}} &
  \textbf{\begin{tabular}[c]{@{}c@{}}$\mathit \Delta$ Site \\ Cover(\%)\end{tabular}} &
  \textbf{\begin{tabular}[c]{@{}c@{}}GN \\ Count(\%)\end{tabular}} \\ \midrule
Amsterdam        & 62.87          & 0.42          & -42.35         & -6.53          & -8.60         & 133.42          \\
Berlin           & 54.60          & 0.38          & -65.77         & -39.84         & -3.64         & 249.49          \\
Munich           & 41.37          & 0.39          & -45.86         & -21.98         & -1.60         & 181.56          \\
Tokyo            & 68.01          & 0.38          & 95.03          & 46.55          & 3.24          & 57.42           \\ \midrule
\textbf{Average} & \textbf{56.71} & \textbf{0.39} & \textbf{62.25} & \textbf{28.73} & \textbf{4.27} & \textbf{155.47} \\ \bottomrule
\end{tabular}

}
\label{tbl3}
\end{table}

The quantitative results in Table~\ref{tbl3} further confirm the model's strong generalization capability. Across the four unknown cities, the model achieved an excellent average FID score of 56.71, showing minimal performance degradation compared to Experiment 1 results on trained cities (average FID 50.94).

Notably, the subtle performance differences across different cities reflect the reasoning process of its knowledge base. For instance, Munich achieved the lowest FID (41.37) and extremely low site coverage error ($\mathit{\Delta} \text{ Site Cover}$ -1.60\%), indicating that its morphology highly matches certain cities in the model's knowledge base. In Tokyo, the model generated significantly fewer buildings (GN Count 57.42\%) than the actual count, potentially reflecting that Tokyo's ultra-high-density unique morphology poses greater challenges to the model's generalization capability, yet its generated overall urban pattern remains reasonable (Figure~\ref{fig7}). Conversely, in Berlin, excessive building generation occurred (GN Count 249.49\%), suggesting the model may have ``misjudged" certain areas of Berlin's morphology as some higher-density style from its knowledge base. This differentiated performance precisely indicates that the model is not mechanically applying templates but rather conducting dynamic morphological reasoning based on its knowledge base.

In summary, whether through controllable style transfer or robust zero-shot generation, both demonstrate that our multimodal model learns flexible, generalizable, and controllable morphological knowledge rather than merely simple image transformation rules.

\subsection{Experiment 3 - Ablation Study on Modality Contributions}\label{4-3}

In this section, we conducted comprehensive ablation experiments on ControlCity to verify the specific contributions of different modality conditions in achieving high-fidelity morphological synthesis. We systematically removed image combinations (road networks and land use), metadata (geographical coordinates), and refined text prompts, then evaluated the resulting performance changes. The quantitative results are presented in Table~\ref{tbl4}, encompassing six key metrics: FID, MIoU, $\lvert \mathit{\Delta} \text{ Bldg. Area} \rvert$, $\lvert \mathit{\Delta} \text{ Bldg. PM} \rvert$, $\lvert \mathit{\Delta} \text{ Site Cover} \rvert$, and GN Count. These metrics represent mean values calculated across ten cities.

\begin{table}[h]
\centering
\caption{Ablation study results showing the impact on performance when removing each modality. Values are averaged across ten cities. Blue indicates the optimal results, while red represents the suboptimal results.}
{
\begin{tabular}{lcccccc}
\toprule
\textbf{Condition} &
  \textbf{FID} &
  \textbf{MIoU} &
  \textbf{\begin{tabular}[c]{@{}c@{}}$\lvert \mathit \Delta$ Bldg. \\ Area$\rvert$(\%)\end{tabular}} &
  \textbf{\begin{tabular}[c]{@{}c@{}}$\lvert \mathit \Delta$ Bldg. \\ PM$\rvert$(\%)\end{tabular}} &
  \textbf{\begin{tabular}[c]{@{}c@{}}$\lvert \mathit \Delta$ Site \\ Cover$\rvert$(\%)\end{tabular}} &
  \textbf{\begin{tabular}[c]{@{}c@{}}GN \\ Count(\%)\end{tabular}} \\ \midrule
{\color[HTML]{000000} w/o Image} &
  {\color[HTML]{000000} 160.72} &
  {\color[HTML]{000000} 0.193} &
  {\color[HTML]{000000} 104.88} &
  {\color[HTML]{000000} 58.61} &
  {\color[HTML]{000000} 12.99} &
  {\color[HTML]{F54A45} 146.25} \\
{\color[HTML]{000000} w/o Text Prompt} &
  {\color[HTML]{000000} 54.28} &
  {\color[HTML]{000000} 0.355} &
  {\color[HTML]{000000} 39.98} &
  {\color[HTML]{000000} 26.26} &
  {\color[HTML]{F54A45} 4.02} &
  {\color[HTML]{000000} 190.90} \\
w/o Metadata &
  {\color[HTML]{4E83FD} 49.63} &
  {\color[HTML]{F54A45} 0.360} &
  {\color[HTML]{4E83FD} 22.60} &
  {\color[HTML]{4E83FD} 11.19} &
  {\color[HTML]{000000} 4.11} &
  {\color[HTML]{000000} 157.88} \\
ControlCity &
  {\color[HTML]{F54A45} 50.94} &
  {\color[HTML]{4E83FD} 0.362} &
  {\color[HTML]{F54A45} 27.00} &
  {\color[HTML]{F54A45} 13.67} &
  {\color[HTML]{4E83FD} 3.82} &
  {\color[HTML]{4E83FD} 145.20} \\ \bottomrule
\end{tabular}
}
\label{tbl4}
\end{table}

\begin{figure*}[!t]
	\centering
	\includegraphics[width=\textwidth]{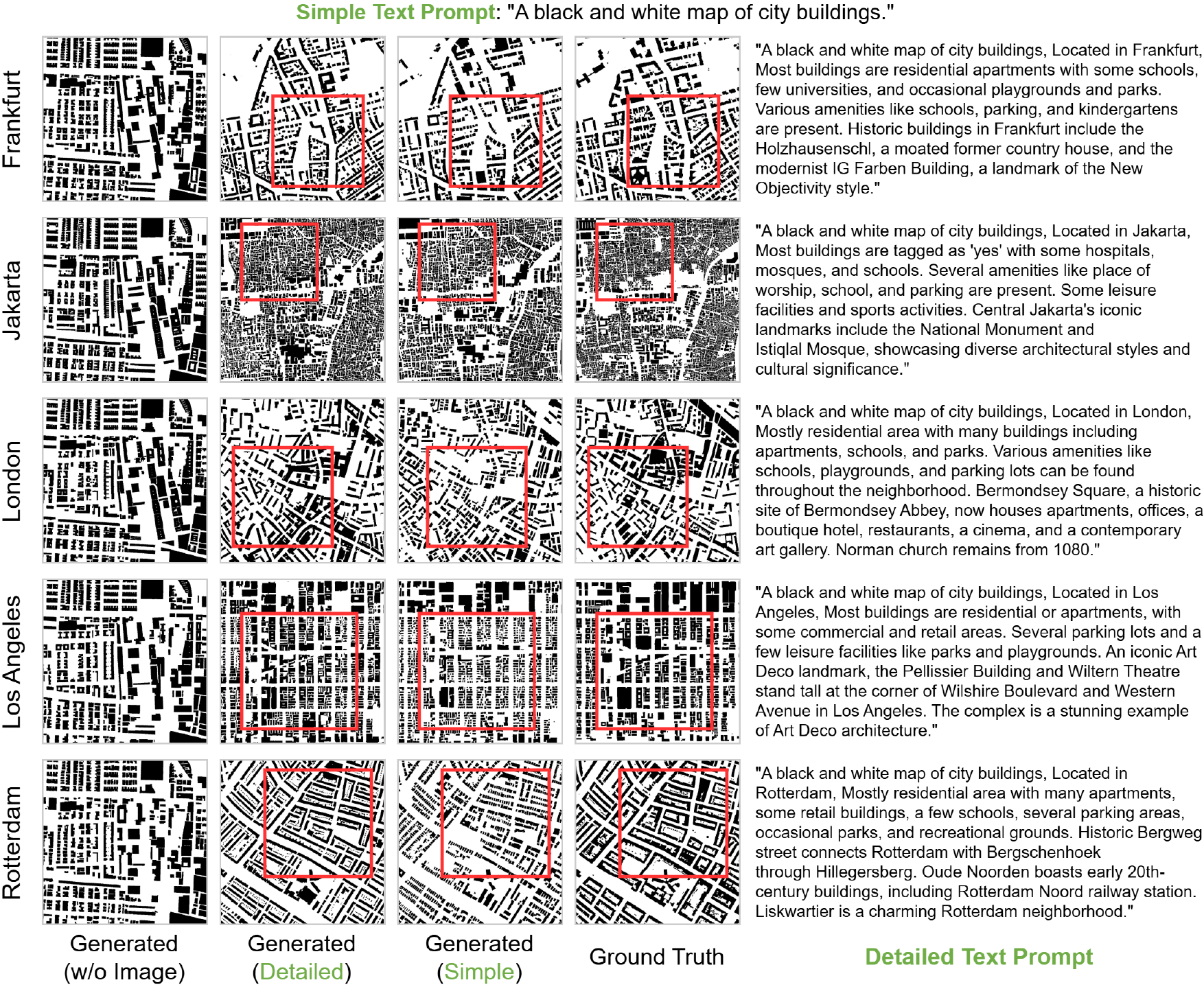}
	\caption{Visualizing the distinct contributions of image and text modalities. The ``Generated (w/o Image)" column demonstrates that removing the image modality causes a complete generative collapse into random texture. Concurrently, a comparison of ``Generated (Detailed)" versus ``Generated (Simple)" reveals that rich semantic input (Detailed Text Prompt) is crucial for morphological accuracy and stylistic control.}
	\label{fig8}
\end{figure*}

\textbf{Effect of Road and Land Use Images.} In ControlCity, the image modality comprises road network and land use image combinations, primarily providing physical structural information for urban morphology. The results in Table~\ref{tbl4} demonstrate that removing image conditions leads to catastrophic performance degradation across all metrics. The fundamental reason for this failure lies in the fact that, lacking the underlying spatial structure provided by the image modality (e.g., street orientations, block boundaries, functional zoning), the model loses the most basic canvas and anchor points for generating building layouts, with output results degenerating into random textures devoid of spatial order (Figure~\ref{fig8}, Generated w/o Image). Relying solely on text and coordinates, which are two highly abstract information sources, the model cannot infer specific building positions and arrangement patterns. This results in spatially disordered and random generation outcomes, consequently causing comprehensive collapse in visual quality (FID 160.72), spatial overlap (MIoU 0.19), and morphological metrics (e.g., $\lvert \mathit{\Delta} \text{ Site Cover} \rvert$ 12.99\%). Interestingly, the GN Count (146.25) remains similar to the complete model (145.20); however, this does not represent effective generation but rather the model producing noise points with coincidentally similar quantities within random textures, lacking any morphological significance.

\textbf{Effect of Text Prompt.} Text prompts provide rich semantic descriptions, enabling the model to better capture specific details and styles in generated building contours. In the text prompt ablation, we replaced refined text descriptions containing urban functions, landmarks, and community characteristics with minimal text descriptions (i.e., ``A black and white map of city buildings."). 

The quantitative results (Table~\ref{tbl4}) indicate that removing refined prompts has minimal impact on visual quality (FID increasing slightly from 50.94 to 54.28) and overlap (MIoU maintaining 0.36), but significantly affects fine-grained morphological control. Most notably, GN Count deteriorates substantially from 145.20 to 190.90, indicating that simple text prompts cause severe over-generation of building quantities. Concurrently, building area and perimeter errors ($\lvert \mathit{\Delta} \text{ Bldg. Area} \rvert$, $\lvert \mathit{\Delta} \text{ Bldg. PM} \rvert$) also increase markedly.

This loss of statistical control manifests as morphological style degradation in the visual comparison shown in Figure~\ref{fig8}. Under simple text prompts, the model can generate basic building layouts but exhibits generalized and chaotic styles. When detailed prompts are provided, the morphological fidelity of generated results improves significantly. For instance, in the Jakarta case, detailed prompts enable the generation of building clusters with more uniform density and consistent style compared to ground truth, avoiding the scattered and irregular building morphologies that emerge under simple prompts. In Frankfurt, detailed prompts more accurately reproduce the distinctive courtyard-style layout formed by large building enclosures, whereas simple prompts result in loose and fragmented courtyard structures. This demonstrates that text prompts serve as the critical modality for injecting ``semantic essence," proving essential for controlling stylistic details in generated results and ensuring reasonable building quantities.

\begin{figure*}[!t]
	\centering
	\includegraphics[width=\textwidth]{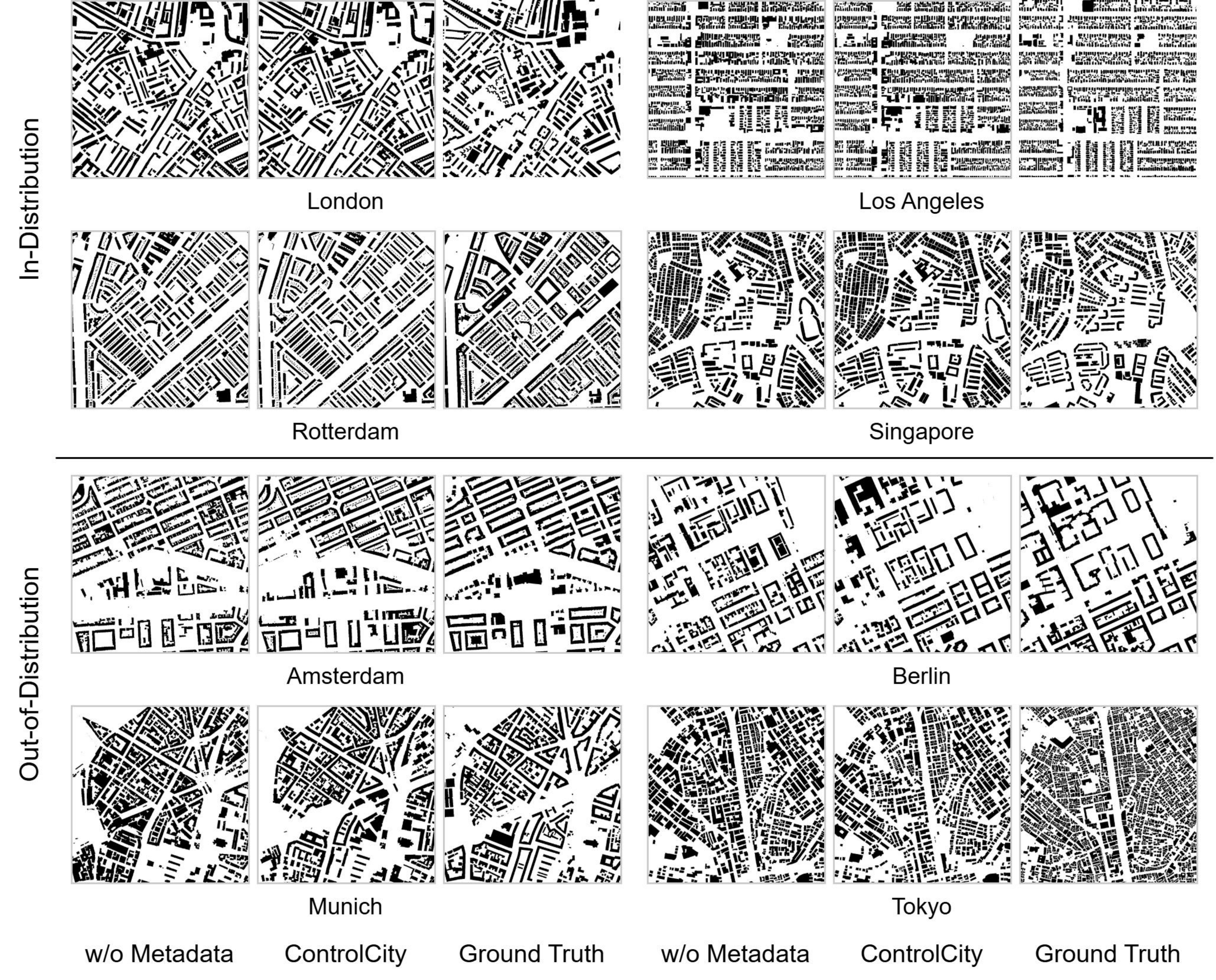}
	\caption{Visualizing the effect of metadata. A comparison of results with and without geographical coordinates for in-distribution and out-of-distribution cities highlights metadata's critical role in zero-shot generalization.}
	\label{fig9}
\end{figure*}

\textbf{Effect of Metadata.} The role of metadata (geographical coordinates) is most nuanced. Table~\ref{tbl4} reveals that after removing metadata, the model achieves sub-optimal performance across multiple metrics: its MIoU (0.36) and morphological errors (e.g., $\lvert \mathit{\Delta} \text{ Bldg. Area} \rvert$ 22.60\%) rank second only to the complete model. This quantitatively confirms that within trained cities (Figure~\ref{fig9}, In-Distribution), when image and text information are sufficient, metadata primarily serves a minor auxiliary calibration role.

However, this delicate balance is disrupted in zero-shot generalization tasks, where the decisive role of metadata becomes prominent. As illustrated in Figure~\ref{fig9} (Out-of-Distribution), when confronting previously unseen European cities, the complete ControlCity model leverages geographical coordinates as prior knowledge to correctly generate courtyard-style layouts characteristic of European urban morphology. In contrast, without geographical coordinate guidance (w/o Metadata), the model exhibits style drift: the generated Amsterdam style erroneously shifts toward American-style rectangular blocks and detached houses; the courtyards in Berlin and Munich become more chaotic, filled with fragmented small buildings.

Notably, removing metadata does not significantly impact overall visual quality (FID score changes from 50.94 to 49.63). However, this macroscopic visual stability masks its negative effects on fine-grained morphological control. The true revelation of its value lies in the GN Count metric, which deteriorates substantially from 145.20 to 157.88, clearly indicating that without global guidance from geographical context, the model's ability to control precise building quantities decreases significantly. Therefore, we conclude that metadata retains irreplaceable value in ensuring statistical accuracy and geographical authenticity of model-generated results.

The results of the ablation experiments reveal a clear and hierarchical collaborative mechanism of modality contributions. First, the image modality (road networks and land use) serves as an indispensable ``physical skeleton," and its removal results in catastrophic performance degradation, demonstrating that spatial constraints constitute the fundamental prerequisite for morphological generation. Second, textual prompts function as the ``semantic essence" infused into the skeleton, responsible for fine-tuning style, functionality, and details. Upon removal of detailed textual information, while macro-level visual quality remains stable, the model's fine-grained morphological control capability is significantly compromised. Finally, metadata (geographic coordinates) provides indispensable ``geographic context," serving as a regional style stabilizer that effectively prevents geographical style drift, ensuring that generated morphologies conform to their macro-regional characteristics. Therefore, the contributions of the three modalities present a synergistic hierarchical relationship characterized as ``skeleton-essence-context": images establish the physical feasibility boundaries of generation; text injects semantics and style within these boundaries; while metadata provides macro-level geographic context for the entire generation process, ensuring the authenticity and rationality of the final results.

\section{Discussion}

\subsection{Sharing of Morphological Knowledge and Model Robustness}

The experimental results clearly demonstrate that the proposed multimodal comprehensive generation model (ControlCity) significantly outperforms unimodal baselines in morphological fidelity. This advantage is manifested not only through quantitative metric improvements but fundamentally stems from its exceptional capability to handle complex and imperfect real-world data. In spatial science research, ideal data environments rarely exist, and a model's true value lies in its robustness when confronting heterogeneous data quality.

A key challenge in this experiment was data quality, where two Chinese cities—Beijing and Shanghai—provided excellent ``stress test" cases. Research by \citet{5-01} indicates that building data completeness in OSM for these cities may be less than 20\%. In Experiment 4.1, the unimodal baseline (Pix2PixHD) encountered complete failure in these two cities, which was unsurprising. Its generation logic relies entirely on single-modality input data; when the input data itself is sparse and incomplete, the model loses effective learning and inference foundations.

\begin{figure*}[!t]
	\centering
	\includegraphics[width=\textwidth]{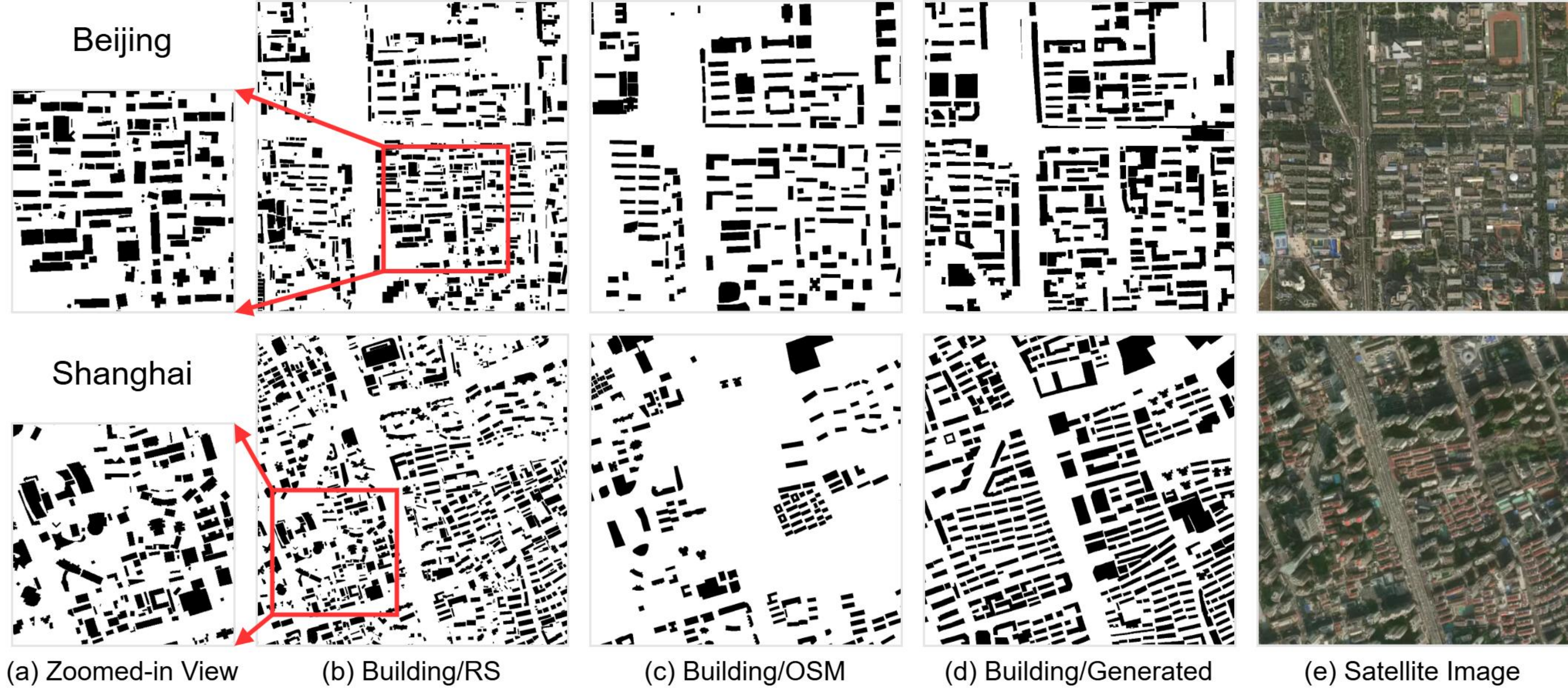}
	\caption{Model robustness to incomplete OSM data. For data-sparse cities like Beijing and Shanghai, the model generates complete layouts (d) that align with satellite imagery (e) by inferring and repairing morphology from incomplete input (c).}
	\label{fig10}
\end{figure*}

However, ControlCity, under the same data conditions, still generated morphologically reasonable building layouts that were far more complete than the input data. As shown in Figure~\ref{fig10}, the model even generated buildings that were missing in OSM but clearly visible in remote sensing imagery (data from \citet{5-02}). This robust performance and generalization capability stems from what we term the ``Morphological Knowledge Sharing" mechanism. Figure~\ref{fig10} provides compelling visual evidence for this mechanism. In Figure~\ref{fig10}(c) OSM data, large areas of Beijing and Shanghai are blank, which is the direct cause of unimodal baseline model failure. However, Figure~\ref{fig10}(d) ControlCity's generation results not only fill these gaps but also show building layouts that are highly consistent with the real conditions presented in Figure~\ref{fig10}(e) satellite imagery in terms of morphology and density. This clearly indicates that the model is not creating from nothing but rather invoking morphological knowledge learned from other complete urban datasets to reasonably ``repair" and ``synthesize" incomplete regions.

This knowledge sharing capability represents the core advantage of the multimodal, multi-city joint training paradigm. Since ControlCity was jointly trained on data from 22 different global cities, it has internalized a rich knowledge base about ``what cities should look like." When confronted with incomplete data from Beijing or Shanghai, it not only relies on current sparse local information but also invokes morphological knowledge from data-complete regions such as Frankfurt and New York City from its knowledge base to reasonably ``fill in" and synthesize missing parts. The Pix2PixHD-based unimodal approach lacks this capability because its ``knowledge" is confined within a single training city, with accuracy entirely constrained by that city's data completeness, lacking cross-regional reference and error correction capabilities.

In summary, ControlCity's robustness does not stem from perfect fitting to a single data source but from its ability to learn, distill, and share universal urban morphological knowledge from diverse multimodal data. This enables it to demonstrate stability and reliability far exceeding traditional methods when confronting ubiquitous data quality issues in the real world. Future research directions should explore how to combine OSM building data with remote sensing extraction data to compensate for building data in incomplete regions, thereby further enhancing model performance.

\subsection{Deconstructing the Multimodal Urban Morphology Synthesis Mechanism}

The core innovation of our proposed ControlCity model does not stem from breakthroughs in individual technologies, but rather lies in its successful construction of a collaborative multimodal comprehensive generation engine. Previous generative modeling approaches typically rely on single information sources, limiting their depth of understanding of complex urban morphologies. ControlCity, however, achieves a paradigm shift from ``geometric imitation" to ``morphological understanding" through the fusion of information from modalities of different natures. The core of this paradigm shift is a profound ``dialogue" between image and text modalities.

First, the image modality provides the indispensable ``physical skeleton". Road networks define the most fundamental topological structure of urban fabric—the orientation of streets and the size and shape of blocks. Land use information provides higher-level functional zoning. Similarly, non-built areas such as water bodies and green spaces delineate the boundaries of building clusters. These two types of image information collectively define the physical feasibility space for building generation, serving as the cornerstone for ensuring that generated results conform to basic geographic reality. As revealed by ablation experiments (Figure~\ref{fig8}, w/o Image column), in the absence of this fundamental spatial structure, the model output completely collapses into random textures devoid of order. Interestingly, while these disordered generated results differ in detail, they macroscopically tend toward a homogenized chaos lacking urban characteristics, which precisely demonstrates the decisive role of road networks as the skeleton of unique urban identity. This spatial disorder ultimately results in catastrophic degradation of key indicators such as visual quality (Table~\ref{tbl4}), proving that without the skeleton, any meaningful morphological generation is impossible. 

However, the skeleton alone is insufficient to constitute a vibrant city; it requires textual prompts to infuse it with ``semantic essence." Unlike the explicit descriptions of ``what you say is what you see" in traditional text-to-image models, the core of our textual prompts lies in their high-level ``implicit instructions." The fixed phrases in prompts (``A black and white map of city buildings.") serve merely as basic visual conventions, while the real challenge lies in enabling the model to understand non-visual information such as city names, functions, and culture. The model cannot directly draw a word called ``Frankfurt," but must understand the complete set of morphological knowledge associated with it. Ablation experiments (Figure~\ref{fig8}) clearly validate the effectiveness of this implicit guidance; after removing these detailed implicit information, while the model can still generate structurally reasonable skeletons, the results become ``form without spirit," exhibiting serious distortions in building quantity and morphological details. This powerfully demonstrates that our model successfully transforms abstract, implicit semantic knowledge into concrete, precise spatial morphologies. 

If images answer the questions of ``what" and ``where" regarding urban morphology, then text addresses the deeper questions of ``why" and ``how." Singular image information remains silent, while text endows it with rich historical, cultural, and individual connotations. The intriguing aspect of this dialogue is that essence cannot exist independently of skeleton, yet the final form of the skeleton is determined by the essence. Figure 8 intuitively presents how this dialogue between image and textual prompts leads to final high-fidelity generation. Isolated image input (e.g., Figure 8, Generated-Simple column, approximating skeleton only) can produce basic layouts but with generic styling; isolated textual input (e.g., Figure~\ref{fig8}, w/o Image column) completely fails to construct spatial arrangements. Only when both modalities work collaboratively (e.g., Figure~\ref{fig8}, Generated-Detailed column) does a ``1+1$\gg$2" synergistic effect emerge. Semantics from text are precisely filled into the spatial constraints provided by images, guiding the model to generate buildings of specific styles within the correct blocks. This transforms the generation process from blind pixel filling to understanding-based, purposeful creation. 

Finally, geographic coordinates, as metadata, provide the model with indispensable geographic context. They position an isolated map tile within the global macro-coordinate system, serving as a regional style stabilizer. As revealed by ablation experiments, geographic coordinates help the model understand a location's macro-regional characteristics. For example, a North American city at coordinates (-122.18, 47.35) is more likely to exhibit architectural morphology similar to a city at (-123.18, 48.35), while differing markedly from a Southeast Asian city at coordinates (103.90, 1.35). This geographic location-based prior knowledge makes the model's generation results more grounded, conforming to the overall character of their regions and effectively preventing style drift phenomena.

In conclusion, ControlCity's success does not depend on any single modality but lies in its organic fusion of physical skeleton (images), semantic essence (text), and geographic context (metadata). It is precisely this synergistic effect of multimodal information that enables our model to transcend simple geometric generation and achieve deep understanding and synthesis of complex urban morphologies.

\section{Conclusion}

This study addresses a fundamental challenge in urban morphology generation: how to advance from simple geometric mimicry to comprehensive generation based on deep understanding. We demonstrate that multimodal information fusion serves as the key driving force for achieving this paradigmatic shift. To this end, we propose the ControlCity model, which successfully bridges the knowledge gaps of ``semantic blindness" and ``contextual neglect" that arose from previous methods' reliance on single information sources. This is accomplished through the synergistic integration of image ``skeletons" that define physical constraints, textual ``essences" that inject functional styles, and metadata ``contexts" that provide regional backgrounds.

This achievement contributes a novel ``understanding-centered" urban spatial generation methodology to the GIScience field. Our approach transcends the traditional perspective that treats geographic data as purely geometric inputs, deconstructing urban morphology into an organic unity of three dimensions: physical structure, semantic connotation, and geographical context. By constructing a framework capable of fusing heterogeneous geographic information, we advance the field from superficial imitation of spatial patterns toward comprehensive simulation of the complex mechanisms underlying geographic phenomena. This work demonstrates the tremendous potential of multimodal generative models in learning, generalizing, and creatively applying geographic knowledge, opening new pathways for GIScience to address and simulate complex human-environment systems.

At the practical level, this method provides urban planners and designers with unprecedented intelligent tools. Through ControlCity, professionals can transcend static layout generation and conduct dynamic scenario simulations that are semantically-based and integrate specific styles and contexts. For instance, different design schemes can be rapidly evaluated through textual instructions (e.g., ``generate a Parisian neighborhood with London courtyard style"), or abstract design guidelines can be transformed into intuitive visualization schemes. This transition from ``passive generation" to ``active creation" provides powerful new tools for rapid scheme iteration, visualization of urban design guidelines, and automated construction of high-fidelity digital twin environments.

Despite the model's significant progress, its performance remains constrained by the quality and coverage of VGI data. This limitation also points toward future research directions: constructing hybrid data-driven generative paradigms. Future work may explore the integration of more comprehensive remotely sensed extracted data with semantically rich OSM data to create higher-quality training datasets. Our prospective supplementary experiment (Appendix~\ref{appendix-b}) has preliminarily validated the effectiveness of this pathway and indicates that it will be crucial for elevating urban morphology generation fidelity to new heights.



\section*{Disclosure statement}

The authors report there are no competing interests to declare.

\section*{Data and codes availability statement}
The data and code that support the findings of this study are available at https://doi.org/10.6084/m9.figshare.30073096. Model checkpoints of this study are available at https://doi.org/10.6084/m9.figshare.30103996.


\section*{Funding}
This work is supported by the Natural Science Foundation of Zhejiang Province [No. ZCLMS26D0105], the National Natural Science Foundation of China [No.32471860,42571529,32571810,32271869], and Deep-time Digital Earth (DDE) Big Science Program.

\bibliographystyle{unsrtnat}
\bibliography{references}  






\appendix

\section{Detailed Text Prompt Example}\label{appendix-d}
See Table~\ref{tbl5}.
\begin{table}[!t]
\centering
\caption{Detailed text prompts used for the cities shown in Figure~\ref{fig4}. These prompts were generated by an LLM (GPT-4o mini) and served as high-level, implicit instructions to guide the morphological generation.}
{
  \begin{tabular}{lp{5.5cm}lp{5.5cm}}
\toprule
\textbf{City} &
  \textbf{Text Prompt} &
  \textbf{City} &
  \textbf{Text Prompt} \\ \midrule
Beijing &
  A black and white map of city buildings, Located in Beijing, Most buildings are residential with some educational institutions, various leisure facilities, and a few sports amenities. Water features include a pond. A mix of modern and historic sites, including Chegongzhuangxi station on Line 6 of Beijing Subway and the restored Zhalan Cemetery, a former Jesuit burial ground with original tombstones. &
  Frankfurt &
  A black and white map of city buildings, Located in Frankfurt, Most buildings are residential apartments, with some detached houses and a few public facilities like schools, hospitals, and parks. Various roof shapes and leisure areas present. Sachsenhausen, located south of the Main river, is known for its historic old town, vibrant nightlife, cider houses, and Museum Embankment. Landmarks include Henninger Turm and Goetheturm. \\
Jakarta &
  A black and white map of city buildings, Located in Jakarta, Mostly industrial landuse with some commercial and residential areas. Many buildings including schools, mosques, clinics, and offices. Several amenities like banks, schools, and places of worship. Historic Jami Kampung Baru Inpak Mosque, with Indian merchant influences, is a cultural heritage site in Jakarta. Located close to Masjid Al-Anshor, it showcases traditional Indonesian architectural style. &
  London &
  A black and white map of city buildings, Located in London, Mostly residential area with some amenities including schools, places of worship, and restaurants. Several buildings, with a variety of types such as apartments, houses, and industrial structures. Historic Leyton Cricket Ground, once home to Essex County Cricket Club, now hosts local matches. Nearby Norlington School and Leyton Midland Road station serve the community in East London. \\
Los Angeles &
  A black and white map of city buildings, Located in LosAngeles, Most buildings are residential with 1355 houses and 78 apartments. Some leisure facilities include 10 swimming pools and 3 pitches, while there are a few tennis courts., Modern architecture and urban landscape with diverse buildings in Los Angeles. &
  New York City &
  A black and white map of city buildings, Located in NewYorkCity, Mostly residential area with numerous buildings, parks, playgrounds, and swimming pools. Abundant footways, sidewalks, and parking spaces, with various amenities like schools and restaurants., Historic Saint Cecilia's Catholic Church in Greenpoint, Brooklyn, features Romanesque Revival architecture. Nearby landmarks include the Humboldt Street station and the Graham Avenue subway station. \\
Rotterdam &
  A black and white map of city buildings, Located in Rotterdam, Mostly residential area with numerous apartments, some gardens, few schools, several bicycle parking spots, occasional parks, and very few swimming pools., The Bergpolderflat, a modern style gallery flat with steel skeleton, designed in 1933 by architect W. de Tijen. Located in the historic Bergpolder neighborhood, part of protected area since 2014. &
  Seattle &
  A black and white map of city buildings, Located in Seattle, Mostly residential area with many houses and apartments, several parks, some wood and beach areas, a few retail buildings, and occasional leisure and social facilities., A popular diving site, Seacrest Cove 2 in West Seattle, offers training for scuba classes with submerged logs and marine life. Nearby North Admiral boasts historic charm and Schmitz Preserve Park. \\
Shanghai &
  A black and white map of city buildings, Located in Shanghai, Mostly residential area with numerous apartments and some commercial buildings. Few leisure facilities like parks and sports amenities. Occasionally schools and hospitals can be found., Jiashan Road station, an interchange station between Lines 9 and 12 of Shanghai Metro, near Zhaojiabang Road and Damuqiao Road. Operational since 2009, featuring modern urban architecture. &
  Singapore &
  A black and white map of city buildings, Located in Singapore, Mostly residential buildings with some schools, leisure facilities like swimming pools and parks. Few sports amenities such as soccer fields and basketball courts. Occasional social facilities and healthcare services., Modern skyscrapers and colonial buildings blend in the central business district, showcasing a mix of contemporary and historic architectural styles. \\ \bottomrule
\end{tabular}
}
\label{tbl5}
\end{table}

\section{Enhancing Performance with Remote Sensing Data}\label{appendix-e}
\begin{figure*}[!th]
	\centering
	\includegraphics[width=\textwidth]{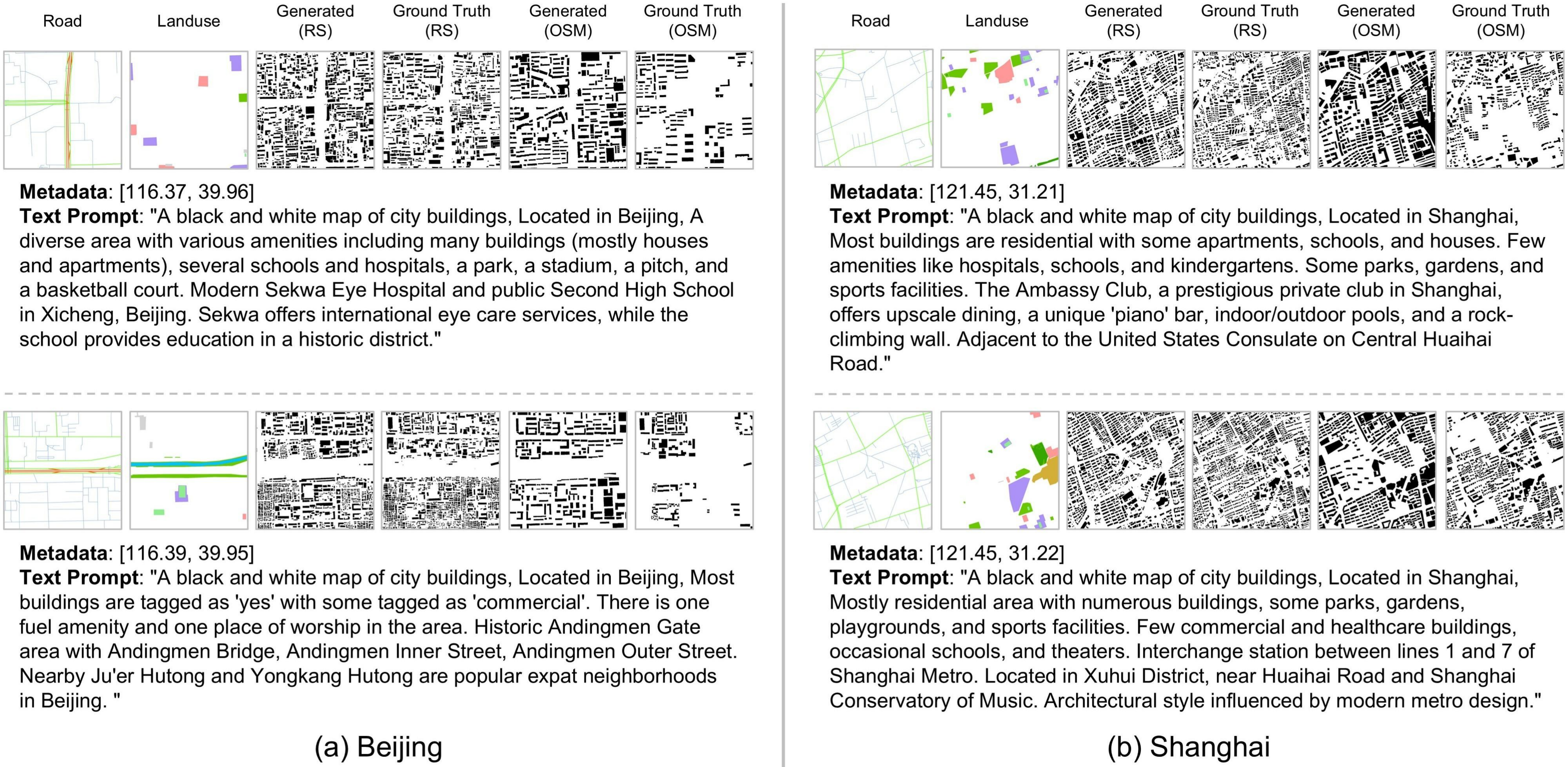}
	\caption{Qualitative comparison of training data sources. The figure contrasts model outputs for Beijing and Shanghai when trained on sparse OSM data versus more complete remote sensing (RS) data.}
	\label{fig11}
\end{figure*}

See Figure~\ref{fig11} and Table~\ref{tbl6}.

\begin{table}[!th]
\centering
\caption{Quantitative comparison of model performance using OSM versus RS data in Beijing and Shanghai.}
{
  \begin{tabular}{cclcccccc}
\toprule
\textbf{Model} &
  \textbf{Data Source} &
  \textbf{City} &
  \textbf{FID} &
  \textbf{MIoU} &
  \textbf{\begin{tabular}[c]{@{}c@{}}$\mathit \Delta$ Bldg. \\ Area(\%)\end{tabular}} &
  \textbf{\begin{tabular}[c]{@{}c@{}}$\mathit \Delta$ Bldg. \\ PM(\%)\end{tabular}} &
  \textbf{\begin{tabular}[c]{@{}c@{}}$\mathit \Delta$ Site \\ Cover(\%)\end{tabular}} &
  \textbf{\begin{tabular}[c]{@{}c@{}}GN \\ Count(\%)\end{tabular}} \\ \midrule
\multirow{4}{*}{ControlCity} & \multirow{2}{*}{OSM} & Beijing  & 55.13 & 0.19 & 20.35 & 15.21 & 8.90 & 142.12 \\
                             &                      & Shanghai & 43.04 & 0.20 & 4.52  & 12.44 & 9.55 & 157.86 \\
                             & \multirow{2}{*}{RS\citep{5-02}}  & Beijing  & 30.56 & 0.19 & -0.59 & -1.11 & 2.40 & 113.69 \\
                             &                      & Shanghai & 45.92 & 0.19 & -0.40 & 1.55  & 3.57 & 126.53 \\ \bottomrule
\end{tabular}

}
\label{tbl6}
\end{table}

\end{document}